\documentclass[12pt]{article}

\RequirePackage{amsmath,amsfonts,amssymb}
\usepackage{multirow}
\usepackage{hhline}
\usepackage[utf8]{inputenc}
\usepackage{url}
\usepackage{graphicx}
\usepackage{times}
\usepackage{cprotect}
\usepackage{lettrine}
\usepackage{textcomp}
\usepackage[dvipsnames]{xcolor}
\usepackage{lineno, blindtext}
\usepackage{tabularray}
\usepackage{adjustbox}
\usepackage{tabularray}
\usepackage{siunitx}
\usepackage{hyperref}
\usepackage{csquotes}
\usepackage{caption}
\usepackage{subcaption}
\usepackage{colortbl}
\usepackage[normalem]{ulem}
\usepackage{soul}

\usepackage{program}

\usepackage[labelsep=period]{caption}
\usepackage[labelfont=bf]{caption}

\definecolor{lightgray}{gray}{0.95}

\title{Testing the Limits of Large Language Models in Debating Humans}

\author
{
James Flamino$^{\text{1,}\dagger}$, Mohammed Shahid Modi$^{\text{1,}\dagger}$,\\Boleslaw K. Szymanski$^{\text{1,*,}\dagger}$, Brendan Cross$^{\text{1}}$, 
Colton Mikolajczyk$^{\text{2}}$\\
\normalsize{$^{\text{1}}$Department of Computer Science and Network Science and Technology Center,}\\
\normalsize{Rensselaer Polytechnic Institute, Troy, NY, USA}\\
\normalsize{$^{\text{2}}$Department of Mathematics, Rensselaer Polytechnic Institute, Troy, NY, USA}\\
\normalsize{$^{\dagger}$Authors contributed equally}\\
\normalsize{$^{*}$Corresponding author, email: szymab@rpi.edu}
}

\date{}

\begin{document} 

\maketitle 

\begin{abstract}  
Large Language Models (LLMs) have shown remarkable promise in communicating with humans. Their potential use as artificial partners with humans in sociological experiments involving conversation is an exciting prospect. But how viable is it? Here, we rigorously test the limits of agents that debate using LLMs in a preregistered study that runs multiple debate-based opinion consensus games. Each game starts with six humans, six agents, or three humans and three agents. We found that agents can blend in and concentrate on a debate's topic better than humans, improving the productivity of all players. Yet, humans perceive agents as less convincing and confident than other humans, and several behavioral metrics of humans and agents we collected deviate measurably from each other. We observed that agents are already decent debaters, but their behavior generates a pattern distinctly different from the human-generated data.
\end{abstract}


\section*{Introduction}

The advent of AI has given rise to a dream of building artificial agents capable of partnering and even successfully competing with humans~\cite{mccarthy2006proposal}. A baseline of successful competition between AI and humans was accomplished in 1997 when IBM's Deep Blue defeated Garry Kasparov, the World Champion at the time~\cite{bloomfield2008ibm}. Later, a notable milestone was set in 2011 with IBM's Watson, which exhibited a robust understanding of human language and defeated human champions in \textit{Jeopardy!}~\cite{ferrucci2012introduction}. In 2022, a breakthrough occurred in AI-human dialogue with the Introduction of Large Language Models (LLMs) such as ChatGPT~\cite{yuan2022wordcraft}. These powerful models, trained on a vast number of human-produced texts, can converse with humans and solve complex text-based tasks~\cite{bubeck2023sparks,argyle2023out,kocon2023chatgpt,shanahan2023role,duenez2023social}. Their ability to realistically interact with people using human language in a contextually and semantically cohesive fashion could transform social science, as postulated in~\cite{grossmann2023ai} wherein the authors discuss scenarios in which LLMs could act as confederates to facilitate human-based experiments or even as surrogates for human players. Both roles have received some attention from designers of LLMs~\cite{argyle2023out,chuang2023simulating,park2024generative}. Although some past work has explored the role of artificial confederates in behavioral experiments~\cite{shirado2017locally}, those confederates acted in a limited role. They lacked the depth of reasoning that LLMs possess.

Testing if LLMs can behave like human confederates is crucial, as these models potentially pose an unprecedented opportunity for researchers to break the traditional mold of human-based experimental design. Today, little is known about the behavioral dynamics of human-LLM conversation, notably in a study setting. Subsequently, in this paper, we describe an experiment that robustly and rigorously tests the limits of agents in acting as confederates or when surrogating humans.

Our paper uses the term 'agents' to describe the simulated players that participate alongside human players in our debate games. These agents generate conversational messages using a combination of two LLMs, GPT-4 and Llama 2.
They use ChatGPT with custom prompts and Python scripts to interact with the game environment and maintain a memory of past conversations within a single game. Thus, any agents with the same persona have no memories of past games, so each game is the first for each agent and human.  The agents are assigned unique 'personas' to individuate them from each other as separate entities. The 'personas' are a unique combination of four initial traits: stubbornness, grammar sophistication, personal confidence and diet. Any agent with the same persona in two or more separate games was considered the same individual in those games. (See "Agent Design" in the Methods section for details on each personality trait and the capabilities of agents.)
 
Within each game, agents, like humans, maintain memories of interactions with other players and could refer to previous discussions in follow-up conversations with players they had already spoken to. Agents did not share memories and were, like humans, unaware that other agents could be present in a game. Each agent cleared memories of the previous games at the beginning of a new game. 
Thus, each game was the first played from the agent's point of view.

We seek to answer two questions relating to the capabilities of these agents in conversation-intensive studies: How capable are agents of acting as confederates (i.e., preserving or enhancing opinion dynamics leading to consensus), and how capable are they of being human surrogates, generating data indistinguishable from human data in the same environment?
 
Considering the implications of these questions, we look to approach this research through the prism of social science. First, we determined the number of samples necessary to achieve statistical significance of our debate experiment's results \cite{lenth2001some, faul2007g}. We identified all the required measures before initiating the study. We then pre-registered the research and implemented a battery of statistical tests (including Bayesian regression models~\cite{kaplan2023bayesian}) on the data collected post-experiment.

The debate experiments we ran consisted of multiple games, starting with six players that involved a combination of humans and agents acting as humans, all anonymously. We had three types of games with seats for either six agents, or six humans, or three agents and three humans, denoted respectively as \textit{AA}, \textit{HH}, and \textit{AH} games. The humans were only informed that agents were engaged at the end of the game. The game aim is to convince other players of your opinion on a debate topic and to seek a majority consensus. For our study, the debate topic was on the best diet choice. Specifically, players were given this prompt at the beginning of a game, verbatim: "Which of these diets is the best compromise between nutritiousness and climate consciousness?" players were then asked to select one of these four choices of diets: Vegan, Vegetarian, Omnivorous, or Pescatarian. We chose this topic for the prompt because a choice in the diet affects a wide range of issues (including environmental and societal) but does not trigger deep societal polarization intrinsic to topics like politics, which often produce opinion stagnation~\cite{mccarty2019polarization}.

To promote collaboration among players, we designed a reward point system: convincing another player to change their opinion grants the persuader one point, and all players with the most popular opinion at the end of the game receive three additional points. At the end of the game, the top two players are awarded double their base compensation. The players are given one hour to debate one-on-one with all other players using a custom texting interface. These rules are revealed to players before the game. The game is intentionally simple in its design, with the sole source of interactions being pairwise, anonymous, open-ended text-based messaging sessions, enabling us to collect all text exchanges, dynamics of opinions about diets, and temporal aspects of interactions. 

Recording the resultant behavioral patterns and comparing them across game types gives us the data to answer our two questions. In the following sections, we describe our experiment and the data we collected, and then we begin analyzing our results and comparing differing behaviors across varied study players.

\section*{Experiment Design}

We deployed our study using voluntary players from the student body of the Rensselaer Polytechnic Institute. A total of $251$ students signed up for the study; 111 humans joined games, but only $N_h=97$ actively participated in conversations. This human attrition led to some AH and HH games having fewer than three or six humans, respectively. We decided to include games with partial populations in the analysis of results since we found that they improve the accuracy of these analyses without skewing the results (see SM section 7, 7.1 and SM table S8). We organized and ran to completion $37$ games, of which $N_{HH}=10$ were \textit{HH} games, $N_{AH}=17$ were \textit{AH} games, and $N_{AA}=10$ were \textit{AA} games. See "Agent design" in the Methods section for the detailed experimental procedure and explaining how we scheduled games and decided human-agent player groupings. 

For the 17 \textit{AH} games, each with three agent seats (51 total seats), we generated 30 agents with unique personas. Some agents with the same personas were present in more than one \textit{AH} game. For the 10 \textit{AA} games with six agent seats each, we generated 60 agents with unique personas. Thus, we had a total of 90 agents with 90 unique personas, and used $N_a = 90$ as the population size of agents.
When analyzing the behavior of agents in the results section, we treat the persona of the agents as an individual subject, employing repeated measures analysis to account for the multiple data points submitted by these subjects. All agents starting a new game are cleared of any memories of the previous games, so all humans and agents play each game for the first time.

A sample size analysis indicated that to achieve a medium effect size in our comparisons, we needed at least $10$ games per type; see the Supplemental Materials (SM) Section 7 for details. We note that for \textit{AH} games, we set the number of games to $17$ to increase the data on human-agent interaction. Player assignment to condition was random; whenever the time slot for a game registered six humans, we would play an \textit{HH} game until we had ten games played. With more than two registered humans, we randomly selected three human players and generated three random agents for an \textit{AH} game. For \textit{AA} games, we generated six random agents.

\begin{figure}[hp!]
\centering
\includegraphics[width=0.9\linewidth]{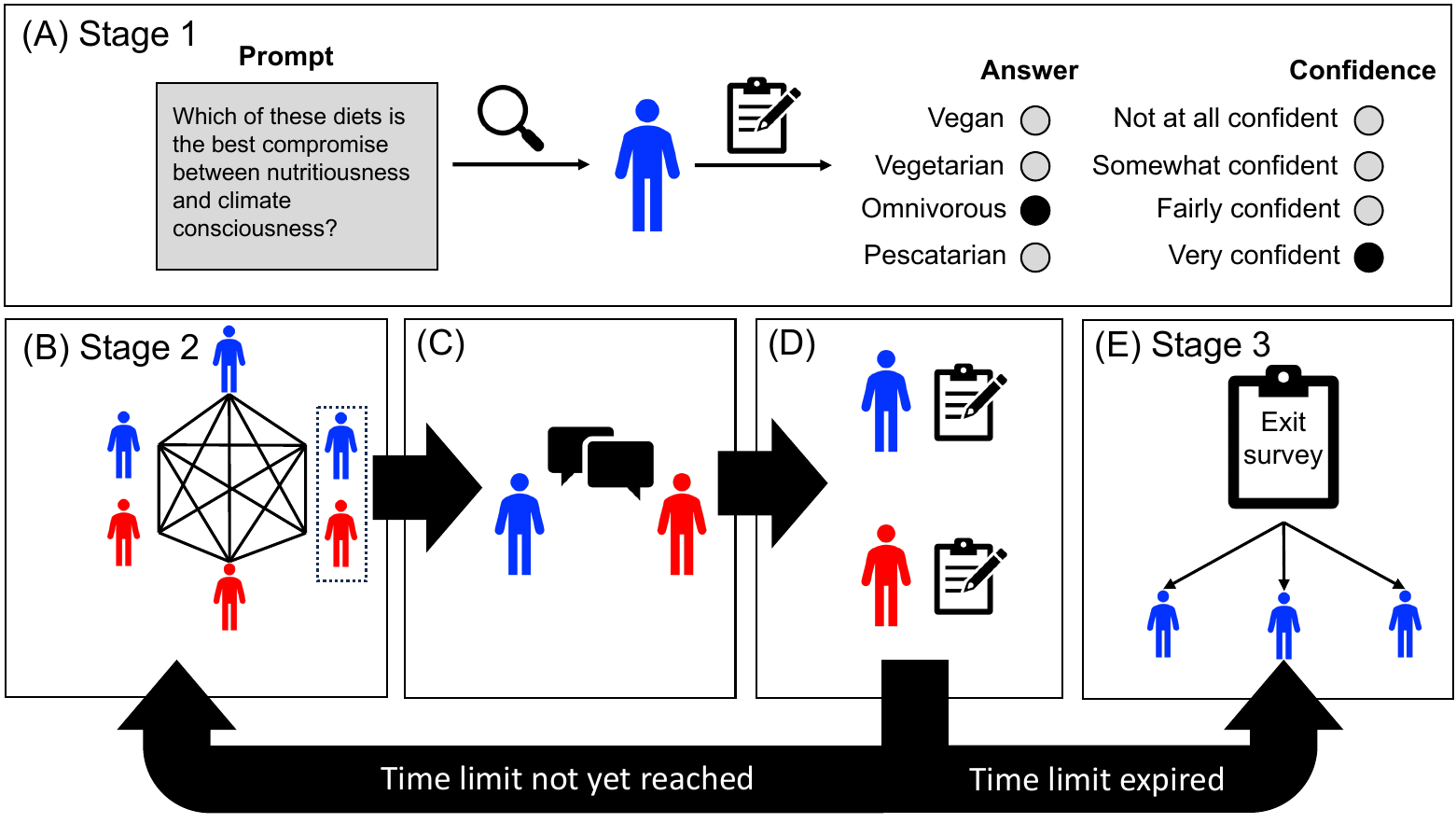}
\caption{\textbf{Experimental setup for a \textit{AH} game type.} (\textbf{A}) Stage 1: The players select an initial opinion and rate their confidence. (\textbf{B}) Stage 2: The players are placed in a fully connected network with five other players who completed stage 1. The \textcolor{blue}{blue} icons represent human players in this network, and \textcolor{red}{red} represent agents. The players can then invite each other to engage in one-on-one conversations. (\textbf{C}) After two players agree to converse, they are cut from the network and converse using a text-based messaging system. (\textbf{D}) Once the conversation terminates, both players re-evaluate their opinions and personal confidence and assign a level of perceived confidence to their conversational partner. Then they rejoin the network (see \textbf{B}). If the time limit has expired, all active players move to stage 3. (\textbf{E}). Stage 3: The game ends, human players file an exit survey, and all players quit the study.}
\label{fig:study_design}
\end{figure}

Figure~\ref{fig:study_design} depicts our game design. In stage 1 of each game (Figure~\ref{fig:study_design}A), each player is given the prompt, which lists four opinion choices from which each player chooses an opinion and rates their confidence in their selection on a four-level scale: "not very confident," "somewhat confident," "quite confident," and "very confident." In stage 2, each player is assigned to a fully connected network of size six. (Figure~\ref{fig:study_design}B). Each player in stage 2 can request a one-on-one conversation with any other player in their network. If their request is accepted, those two players temporarily exit the network and engage in a private conversation via text-based messaging. The conversation lasts until one of the conversational partners terminates it or the time limit expires (Figure~\ref{fig:study_design}C). When a conversation ends, both players can re-evaluate their opinion and their level of personal confidence. They are also asked to rate their partner's perceived confidence using one of the same choices as personal confidence, with an additional "not enough info" option. After completing this step, time permitting, both players are returned to the network, and each can request or accept subsequent conversations from other players (Figure~\ref{fig:study_design}D).

Once the time limit has expired, all active players move to stage 3 (Figure~\ref{fig:study_design}E), in which human players file exit surveys (see "Exit survey design" in the Methods for more details). Then, all players of this game are removed from the game and the study.

Our final dataset includes $713$ conversations, of which $591$ were completed before the time limit expired.$15,804$ messages were sent in those conversations, resulting in $238$ opinion changes. We also collected $91$ exit surveys from the human players in the \textit{AA} and \textit{AH} games. Of all the conversations made in the \textit{AH} games, $152$ were between a human and an agent, $30$ between two humans, and $82$ between two agents, respectively. In the \textit{HH} games, $140$ conversations were between two humans, while in the \textit{AA} games, $309$ conversations were between two agents. Excerpts of conversations occurring in three game types are shown in SM Table~S24.

\section*{Results}

We focus on three behavioral metrics for each player. The first metric is the fraction of conversations that resulted in opinion change. The second is the frequency of changing all self-reported personal and perceived confidences. Our final measurements focus on productivity, which applies metrics of player outputs based on statistical data collected in-game and synthesized post-game. The former involves the number of conversations, their message counts, and the count of the points awarded to players at the end of the games. The latter focuses on conversational attributes, including the fraction of on-topic messages sent during conversations. Both in-game and post-game productivity measures enabled us to detect when humans and agents behave differently comparatively.

\begin{table}[ht]
    \centering
    \resizebox{.9\columnwidth}{!}{
    \begin{tblr}{
      width = \linewidth,
      colspec = {|Q[173]Q[283]Q[237]Q[240]|},
      row{1} = {c},
      cell{1}{1} = {r=2}{},
      cell{1}{2} = {r=2}{},
      cell{1}{3} = {c=2}{0.477\linewidth},
      cell{3}{3} = {r},
      cell{3}{4} = {r},
      cell{4}{3} = {r},
      cell{4}{4} = {r},
      cell{5}{1} = {r=4}{},
      cell{5}{3} = {r},
      cell{5}{4} = {r},
      cell{6}{3} = {r},
      cell{6}{4} = {r},
      cell{7}{3} = {r},
      cell{7}{4} = {r},
      cell{8}{3} = {r},
      cell{8}{4} = {r},
      cell{9}{3} = {r},
      cell{9}{4} = {r},
      vline{2-3} = {1-2}{},
      vline{4} = {2}{},
      vline{2-4} = {3-8}{},
      vline{3-4} = {6-8}{},
      hline{1,3-5,9} = {-}{},
      hline{2} = {3-4}{},
      vlines,
    }
    \textbf{Game Type} & \textbf{Conversation type} & \textbf{Did the player change opinion?} & \\
     &  & \textbf{Yes} & \textbf{No}\\
    \textit{HH} & \textit{hh} & 32 & 218\\
    \textit{AA} & \textit{aa} & 114 & 458\\
    \textit{AH} & \textit{hh} & 7 & 38\\
     & \textit{ha} & 4 & 130\\
     & \textit{ah} & 43 & 91\\
     & \textit{aa} & 38 & 110 \\
     \textbf{Total}&  & 238 & 1046 \\ \hline
    \end{tblr}
    }
    \caption{\textbf{Summary of conversations and opinion change in all games.} This table contains the opinion change outcome of all 642 completed conversations across all three game types. Conversation type indicates the type of user interaction: two humans (\textit{hh}), two agents (\textit{aa}), or an agent and a human (\textit{ah}). For \textit{ah} conversations, we show two rows where the first letter indicates for which player, agent, or human we measure opinion change. Each player in a conversation has the opportunity to change their opinion, yielding $642 \times 2$ opinion changes. We note that incomplete conversations do not record any opinion changes, and there were cases in which one player finished re-evaluating their opinion before the time limit expired, but their conversational partner did not. See SM Table S22 for a version of this table further split by the type of player who initiated the conversation.}
    \label{tab:opinion_change}
\end{table}

With two player types (human and agent) and three game types (\textit{AA}, \textit{AH}, and \textit{HH}), there is a variety of data that can be grouped differently. Subsequently, we classify conversations into three categories: \textit{aa}, \textit{ah}, and \textit{hh}, indicating conversations between two agents, an agent and a human, and two humans, respectively. The homogeneous games can only generate corresponding homogeneous discussions (e.g., \textit{HH} games create \textit{hh} conversations). However, \textit{AH} games can produce all conversation types. When reporting on conversations in \textit{AH} games that involve opinion change, we indicate which type of players changed their opinion by changing the order of the a and h, where the first letter is the indicator. Hence, the behavior of humans in AH games. Each AH conversation generates two data points, one for \textit{ah} and one for \textit{ha}.

A similar distinction is needed when players assign a level of perceived confidence to their conversational partner. As such, we draw an arrow from the assigning player type to receiving player type: \textit{a}$\rightarrow$\textit{a}, \textit{a}$\rightarrow$\textit{h}, \textit{h}$\rightarrow$\textit{a}, \textit{h}$\rightarrow$\textit{h}. These indicate the assignment of perceived confidence from an agent to an agent, an agent to a human, a human to an agent, and a human to a human, respectively. We refer to this notation as "assignment type."

\subsection*{Dynamics of opinion switching}

In consensus games, success is to convince another player to adopt one's opinion. We analyzed how often human players change their opinions when conversing with other humans or agents. We measured the opinion change frequency of players using a Bayesian multilevel logistic regression model predicting the probabilities of such changes as a function of the player and their conversational partner types. See Table~\ref{tab:opinion_change} for all instances of post-conversation opinion changes across the three types of games, clustered by conversation type.

\begin{table}[htp!]
 \centering
 \begin{adjustbox}{width=.9\columnwidth,center}
 \begin{tabular}{|rrrr|}
   \hline
   Conversation type & Odds ratio  & Odds & CI (95\%)  \\
   \hline
   \textit{hh} & 1 & 0.12 & 0.068 - 0.185 \\
   \textit{ha} & 0.16 & 0.02 & 0.003 - 0.046 \\
   \textit{ah} & 2.99 & 0.36 & 0.203 - 0.574 \\
   \textit{aa} & 1.97 & 0.23 & 0.162 - 0.319 \\
    \hline
 \end{tabular}
 \begin{tabular}{|l|l|}
  	\hline
 	  Observations & 1284 \\
     	$N_{game}$ & 37 \\
     	$N_{player}$ & 185 \\
     	$\sigma^2$ & 3.29 \\
     	$\tau_{00\ game}$ & 0.10 \\
     	$\tau_{00\ player}$ & 0.76 \\
     	Marginal $R^2$ & 0.030 \\
     	Conditional $R^2$ & 0.140 \\
   	\hline
   	\end{tabular}
  \end{adjustbox}
  \caption{\textbf{Bayesian multilevel opinion changing model.}
 This table outlines the results of our regression model, presented as the odds of opinion transitions in conversations and odds ratios relative to \textit{hh} conversations. Note that in \textit{ah} and \textit{ha} conversation types, the first letter indicates the player type, which changes opinion. This analysis only includes data from completed conversations. Thus, this analysis excludes the 14 players who didn't actively participate in their respective games and the two who conversed for the entire game time without closing the conversation.}
  \label{tab:bayes_regression_table}
 \end{table}

Table~\ref{tab:bayes_regression_table} summarizes the results of our Bayesian model, shown as the odds of opinion change for humans and agents in each conversation type. The odds ratio compares the odds of opinion change for each conversation type against the human-human (\textit{hh}) conversations, for example Odds-Ratio$_{\text{\textit{ha}}}=$Odds$_{\text{\textit{ha}}} / $Odds$_{\text{\textit{hh}}}$. The 95\% credible intervals (CI) shown are in Odds of opinion change. For more details on the regression model, including formula and raw coefficients, see SM Section 9 and SM Table S11.

\begin{figure}[htp!]
    \centering
    \includegraphics[width=.7\linewidth]{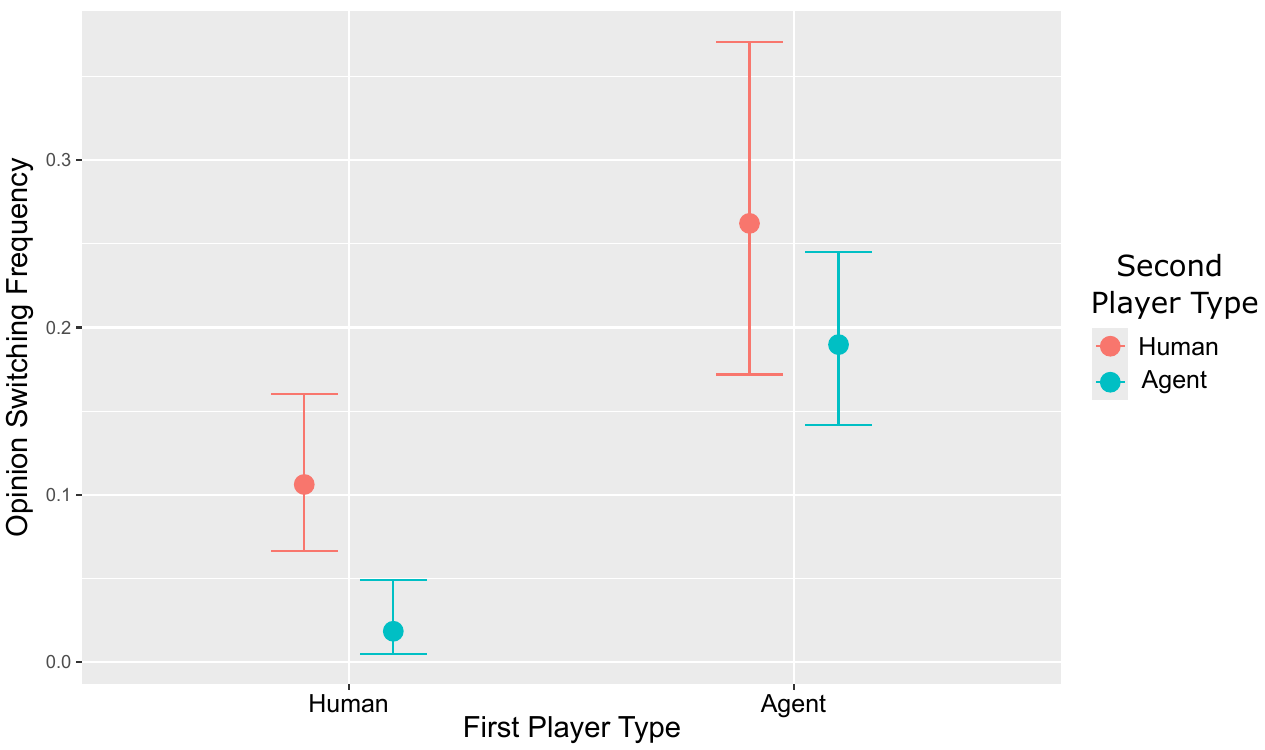}
    \caption{\textbf{Bayesian multilevel model for opinion changing by player type.} This plot shows the frequency of opinion changes as a function of the two conversing players. Each player in a conversation is a human or an agent, and combining these two types gives us an opinion-switching frequency for all conversation types in the study.}
    \label{fig:simple_interaction_model}
\end{figure}

Figure~\ref{fig:simple_interaction_model} shows the mean and credible interval of the opinion-switching frequency for each type of conversation that could occur in the study. In \textit{hh} conversations, the expectation is that $0.12$ players will switch their opinion for every non-changing player mean odds: $0.12$, $95\%$ CI: $0.07 - 0.19$). The \textit{ha} odds ratio (mean value $0.16$, $95\%$ CI: $0.04 - 0.47$) shows that for every opinion switched in \textit{hh} conversations only $0.16$ opinions switch for \textit{ha} conversations. The inverse of this, $6.25$, shows that humans switch their opinions $6.25$ times more often when interacting with other humans vs agents. Therefore, on average, agents need $6.25$ times more conversations with humans to induce the same number of opinion switches that \textit{hh} conversations induce. We also found that the rate of opinion change for \textit{ah} conversations can depend on what type of player holds the opinion proposed for switching to. For more details on this finding, see SM Section~13 and SM Tables~S20, S21, and S22.

Overall, the above results imply for question 1 that agents are far less convincing for humans than humans. Relevant to question 2, the differences in the frequencies of opinion switching seen between agents and humans are confirmed by our regression model in Table~\ref{tab:bayes_regression_table}. When we average over levels of the second player type, we get that humans change opinion with an odds of $0.047$ ($95\%$ CI: $0.023 - 0.083$), while agents change opinion with an odds of $0.288$ ($95\%$ CI: $0.201 - 0.408$). This means agents, on average, have a $6.1$ times higher odds of switching their opinion when compared to human participants.

\subsection*{Personal and perceived confidence}

\begin{figure}
    \centering
    \includegraphics[width=\linewidth]{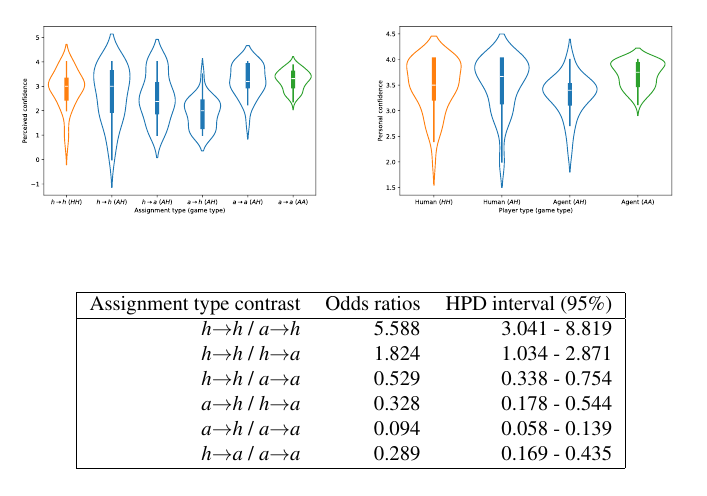}
    \caption{\textbf{Distributions and modeled assignment type contrast for perceived and personal confidences.} (\textbf{A}) This shows violin plots of perceived confidence, clustered by type of assignment and of game (indicated in parentheses). (\textbf{B}) This shows violin plots for personal confidence, clustered by player type and game type (also indicated in parentheses).
    (\textbf{C}) The embedded table shows the contrasts between assignment types in the perceived confidence Bayesian regression model. Estimates are given as odds ratios between contrasting conversation types. The highest posterior density (HPD) interval defines the shortest interval containing 95\% of the posterior mass for the given estimates.
    }
    \label{fig:perceived_confidence}
\end{figure}

Figure~\ref{fig:perceived_confidence} displays the distributions of confidence exhibited by players by type of game and assignment for both perceived and personal confidence. We numerically encoded the confidence levels, from the "Not very confident" level assigned as 1 to "Very confident" assigned as 4. Additionally, "Not enough info" was assigned 0 for perceived confidence. In Figure~\ref{fig:perceived_confidence}\textbf{A}, we clustered perceived confidence levels by their assignment type, while for Figure~\ref{fig:perceived_confidence}\textbf{B}, we clustered personal confidence levels by player type.

We again use a Bayesian multilevel model to compare the distributions of perceived confidence from \textit{h}$\rightarrow$\textit{h} and  \textit{h}$\rightarrow$\textit{a}. The model reveals a statistically significant difference in how perceived confidence is assigned. Furthermore, the model suggests a substantial difference in the perceived confidence being assigned by humans to agents compared to agents to humans. The complete model specifications are shown in SM Section~9 and SM Table~S12.

In terms of results, Table~\ref{fig:perceived_confidence}\textbf{C} shows the assignment type contrasts in perceived confidence assignment for the perceived confidence model. The model establishes that humans perceive other humans as more confident than agents by mean value $1.8$, $95\%$ HPD-interval: $1.034 - 2.871$). Conversely, agents assign a significantly lower level of perceived confidence when interacting with humans compared to other agents by a factor of $10$ (mean value $0.094$, $95\%$ HPD-interval: $0.058 - 0.139$). These differences in perceived confidence between humans and agents are statistically significant for most assignment types.

We extend this analysis to personal confidence as well. Implementing a Bayesian Multilevel model to predict personal confidence as a function of conversation type, we found that no statistically significant relation exists between conversation type and the players' confidence (see SM Table~S13). Concerning question 1, the above analysis indicates that agents negatively affect a human's perceived confidence but do not affect personal confidence.

The above results have implications for question 2 as well. Notably, agents consistently perceive other agents as more confident, and humans tend to treat other humans likewise. The consistency of this behavior identifies a degree of similarity in the two-player types in terms of assigning perceived confidence to the same player type.

\subsection*{Post-game productivity}

\begin{figure}
\centering
\includegraphics[width=0.74\linewidth]{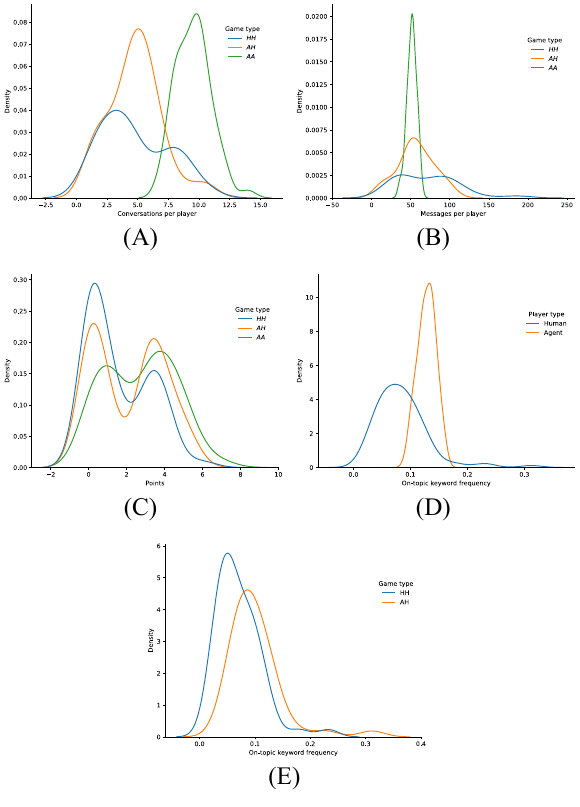}
\caption{\textbf{Post-game and in-game productivity results.} (\textbf{A}) Kernel Density Estimations (KDEs) of the number of conversations per player grouped by the type of game they played in. (\textbf{B}) KDEs of the number of messages each player sent within each game, grouped by their type. (\textbf{C}) KDEs of the reward point distributions gained by all players at the end of each game, grouped by their type. (\textbf{D}) KDEs of the on-topic keyword frequency of each player, agent and human, across all game types. (\textbf{E}) KDEs of the on-topic keyword frequency of each human in \textit{AH} game versus humans in \textit{HH} games.
}
\label{fig:post_game_productivity}
\end{figure}

Next, we analyzed changes in productivity. The first three sub-figures of Figure~\ref{fig:post_game_productivity} show the results of post-game analyses on the number of conversations and message counts (Figure~\ref{fig:post_game_productivity}\textbf{A} and \textbf{B}) and point distribution (Figure~\ref{fig:post_game_productivity}\textbf{C}).These measures are aggregated over each player (human participant and agent persona) such that each player provides one data point representing their average behavior throughout the study. For the conversation and message counts, our visualizations make apparent a spectrum of behavior in \textit{HH} games and \textit{AA} games. Conversations and messages from \textit{AA} games tend to exhibit distributions with a tighter modality, indicating that agents' communication behavior is much more static. This makes sense, as the agents were designed with a message budget: a maximum number of messages allowed to be sent before the conversation is terminated. See "Agent design" in the Methods for details on the implementation. We also further explore the relationship between the number of messages sent per conversation and the influence of the message budget in SM Section~8.

Compared to the communication behavior seen in \textit{AA} games, \textit{HH} communication varies much more, as illustrated in the wider modalities of Figure~\ref{fig:post_game_productivity}\textbf{A} and \textbf{B}. The \textit{AH} games show a mediation of these two extremes, indicating the presence of agents in these games adds a level of predictability. In the context of question 1, we found that the more agents narrow their variances of conversation length and count, the more humans conversing with these agents narrow their corresponding variances too.

Agents also facilitated the distribution of award points in games, earning more points overall than humans (Figure~\ref{fig:post_game_productivity}\textbf{C}). This suggests that agents were likelier to stay focused and attempt to promote opinion change in themselves and others. This observation is essential for question 2, indicating that agents are more goal-oriented in this setting. We report majority consensus groups for \textit{AH} games in SM section 15 and SM Table~S23.

\subsection*{In-game productivity}
The second subcategory in our productivity analysis relates to in-game productivity, like on-topic keyword usage. Here, we define on-topic keyword usage as the number of occurrences of specific words that we identified as related to the game's prompt (i.e., diets, nutrition, climate) or opinion consensus (i.e., team, majority, minority). More information on the dictionary of keywords used can be found in "Keyword analysis" in the Methods.

Figure~\ref{fig:post_game_productivity}\textbf{D} shows KDEs of the frequency of on-topic keywords for each player involved in the study, grouped by their player type (i.e., an agent or a human). Each player provided one data point representing their frequency of using on-topic keywords across all conversations they participated in. One primary observation from this data is that agents generate more on-topic keywords than humans. The mean and standard deviation of the human distribution from Figure~\ref{fig:post_game_productivity}\textbf{D} are $0.085$ and $0.046$, respectively, compared to $0.129$ and $0.016$ for the agent distributions.

Figure~\ref{fig:post_game_productivity}\textbf{E} presents a similar analysis to figure~\ref{fig:post_game_productivity}\textbf{D}, except we only consider human player on-topic keyword frequencies and group them by the type of game they participated in. Importantly, we find that agents' proclivity for on-topic discussion affects the behavior of humans that are in the same game, as indicated by humans in \textit{AH} games generate more on-topic keywords than humans in \textit{HH} games ($p=$ 0.0018, Welch's t-test~\cite{ruxton2006unequal}). We extend this analysis to unique word counts in SM Section~10. Additionally, we further probe the dynamics of player behavior during conversation by analyzing messages' response times and holding periods, shown in SM Section~2 and SM Table S3. This analysis reinforces that agent presence affects humans during conversation, with agents often making humans slower to respond. We also probe changes in in-game productivity based on conversation initiator in SM Section~13 and SM Tables~S17, S18, and S19.

Ultimately, these results shed further light on question 1, as agents can elevate human engagement. Furthermore, in answer to question 2, the results demonstrate that agent word usage during conversation is measurably different from humans.

\subsection*{Exit survey results}

In stage 3 of our game (Figure~\ref{fig:study_design}E), all human players are asked to fill out an exit survey, in which they must nominate the players they thought were the most and least convincing. At the end of the study, for the \textit{AH} games, $18$ agents and $24$ humans were nominated as the most convincing players. In contrast, $31$ agents and $11$ humans were nominated as the least believable. These results indicate that the human players who took the exit survey were likelier to nominate other humans than agents as most convincing ($p=0.0025$, Boschloo's exact test).

\subsection*{Agents' ability to blend in}

Considering the tenuous relationship forming between humans and AI in the face of the fast-paced integration of AI into everyday life~\cite{baronchelli2024shaping, federspiel2023threats}, we implemented an exploratory measure for the agents' ability to blend in with the human players during the games. We kept track of the instances in which humans accused a player of being an agent, which we call an agent detection incident (ADI). To recognize ADIs, we implemented a keyword filter on all messages produced by humans in the \textit{AH} game type. This filter flagged any messages where the words "bot", "AI", "ChatGPT", or "chatbot" were mentioned. Within \textit{AH} games, $14$ human players identified an agent during the conversation. This is $27.4\%$ of the human player population in \textit{AH} games, with these ADIs occurring across $9$ of the games, leaving $8$ \textit{AH} games with no incidents. In the $9$ \textit{AH} games with ADIs, $13$ of the human players did not identify any agents. Therefore, humans who detected an agent did not always spread that information to the other human players. Interestingly, we observed that agents failing to cover their identity during one conversation does not imply that they will continue to fail afterward. One false ADI was also in an \textit{HH} game.

 \newcolumntype{g}{>{\columncolor{lightgray}}c}
 \begin{table}[t!]
   	\begin{center}
         	\begin{tabular}{|p{0.5cm}|p{14cm}|}
               	\hline
                 \rowcolor{lightgray}
               	\textbf{$\#$} & \textbf{Cause of ADI} \\
               	\hline
               	1 & System: Agent cut the conversations abruptly when their message budget is empty.\\
               	\hline
               	2 & System: Agent responded to itself. \\
               	\hline
               	3 & System: The agent greeted the user in the middle of the conversation multiple times.\\
               	\hline
               	3 & System: The agent greeted the user during the conversation.\\
               	\hline
               	3 & System: The agent got confused about the past conversation in the follow-up conversation. \\
               	\hline
               	4 & System: The agent responded to itself and argued against itself. \\
               	\hline
               	4 & Language: Agent repeated its phrasing too closely. \\
               	\hline
               	5 & Language: The agent was too lengthy, and there was a delay in its replies. \\
               	\hline
               	4 & Language: Agent repeated its phrasing too closely. \\
               	\hline
         	3 & Language: The agent was too lengthy and formal. \\
               	\hline
               	1 & Language: The agent was too lengthy and often triggered the not-on-topic flag. \\
               	\hline
               	6 & Language: The agent was too formal and repeated its Introduction later. \\
               	\hline
         	7 & Language: The agent was too lengthy and confident. \\
               	\hline
         	\end{tabular}
   	\end{center}
   	\caption{\textbf{\textit{ah} conversation ADI causes.} This table shows 13 \textit{ah} conversations with Agent Detection Incidents that were system or language-provoked. The game number (first column) indicates the game where the conversation occurred, with some games having more than one conversation. Eight \textit{ah} conversations with humans "hunting" for agents are not listed, and six \textit{hh} conversations with human-human spread ADIs.}
   	\label{tab:adi_annotations}
 \end{table}

We found that not enough players were involved to statistically verify any change in human behavior in the presence of ADIs (see SM Section~4 and SM Tables~S4, S5, S6, and S7 for more details). However, we probed for more subtle behavioral changes by analyzing conversational partner selection bias in \textit{AH} games (see SM Section~11). This test established that humans did not prefer connecting to other humans over agents in-game, and the same behavior was also found in the agents. We also performed a thorough qualitative study of why agents were detected and their context. We analyzed every ADI from the \textit{AH} games to see what agent and human behavior led to agent detection. We designated four "triggers of detection" categories where at least one can be assigned to each ADI case. These categories are \textit{AI system provoked}, \textit{AI language provoked}, \textit{human provoked}, and \textit{human-human spread}. The categorization methodology is described in "Agent detection categories" in the Methods section. There were 21 \textit{ah} conversations and 7 \textit{hh} conversations with ADIs. Of those 21 \textit{ah} conversation ADIs, 13 were caused by AI system/language provoked detections (61.91\%), and the other 8 had human "hunters" with pre-existing suspicion. Table \ref{tab:adi_annotations} shows agent-provoked ADI causes. It follows that  13 of 21 humans in ADI \textit{ah} conversations could have found agents organically and not from prior suspicion.

\section*{Discussion}

In the Introduction, we established two questions we sought to answer with our study: how capable are agents in preserving or enhancing opinion dynamics leading to consensus, and whether agents can generate realistic, human-like debate data. By collecting data on human and agent interactions in opinion consensus games where the nature of the players is anonymous, our results shed some light on the behavioral differences between agents and humans and the nature of their interactions.

While it is important to note that our samples are not representative of the general population, it can be argued that the study's population tests the limits of the AI agent's ability to blend in, as the younger, more educated demographic of our research may be more capable of identifying patterns of AI behavior. As such, our results should be considered as part of stress-testing conversational AI capabilities instead of simply being a representative snapshot of expected behavior for humans in the US. In this context, our analysis can be considered more generalizable to the nascent and trending social media platforms~\cite{Zhou_2024} where AI technology has or most certainly will be integrated.

In the rigorous analysis of our study's data, we found that agents attempting to influence people to change their opinions are six times less likely to succeed than humans influencing other humans. This trend persists with perceived confidence assignment since humans perceived other humans as $1.8$ times more confident than agents. Agents were found to switch opinions 6.1 times more often than human players. Agents also found other agents to be $10$ times more confident than humans (Table~\ref{fig:perceived_confidence}\textbf{C}). Furthermore, agents produced more on-topic keywords when in conversation, even elevating the usage of on-topic keywords for humans that participated in games with them (Figure~\ref{fig:post_game_productivity}). 

An interesting observation from the above results is that agents' productivity in initiating conversations and staying on-topic resulted in an increased number of points awarded among agents despite the agents' less convincing and confident behavior. This outcome suggests that many agents initiated conversations and provided compelling arguments for their diets. This strategy is effective in a debate-based opinion consensus game.

In Summary, while our agents can increase the productivity of humans participating in an opinion consensus game, they hinder humans' opinion-changing behavior and appear less confident in conversation. Subsequently, they must improve in this area to be capable confederates. Our agents' underlying behavior fundamentally differs from humans'. They stay on-topic longer than humans. These differences in behavior set them apart from humans, so much so that we can use simple machine learning classifiers trained on a subset of the metrics discussed above to differentiate between \textit{ah} and \textit{hh} conversations and \textit{AA} and \textit{HH} games (see SM Section~12 and SM Tables~S14, S15, and S16). This deviation in behavior means that our agents are not yet ready to become accurate human surrogates in producing behavioral data. 

One limitation of our work is that it may not reflect how humans with prior knowledge of agent presence may behave in debates and conversations. Our goal was to investigate if agents can be effective confederates in such contexts. If we labeled agent players as agents from the beginning, humans may have approached them differently, even when sharing the same goal of reaching a consensus. We chose not to suggest anything about who the players are to avoid discounting agents' arguments in the debate. Such a bias of humans would distort all measures of the agent's capabilities to debate. We were also careful not to suggest that players are all humans by using colors to identify players instead of giving agents human names. The results indicate that initial good agent performance in the games in which a human player detected agents' presence gave agents a chance to gain enough respect from humans that they stayed in the games.
Moreover, the interrupted games were statistically similar to the uninterrupted games. Another limitation is the unknown generalizability of our results to AI generative models other than the ones we used. All current studies share this limitation since LLM models rapidly evolve. We expect that the initial uneasiness to accept agents some humans have will quickly dissipate as interactions with agents become ubiquitous, making our approach neutral to results.

We conclude that our agents show promise in conversing with humans but must evolve more before becoming capable surrogates or confederates in conversation-intensive sociological studies. Future work is essential, given that research on human-AI dynamics within sociology remains unexplored. Our paper creates a foundation for investigating interactions between humans and agents in a debate setting. In sociological studies, it establishes an archetype for exploring these or other LLM-based confederates. In future research, we plan to study the relationship between AI language more deeply, how humans discover AI in anonymous interactions, and semantic drift~\cite{spataru2024know}.

\section*{Methods}

Due to the nature of the study, this research was reviewed and classified as exempt by the Rensselaer Polytechnic Institute (RPI) Integrated Institutional Review Boards (IRB). This decision is shown in IRB ID 2133 (approved on 5 September 2023). These files are available upon request of the corresponding author. Executing all methods, we applied the relevant guidelines and regulation. All players confirmed informed consent before entering the study. Our study's experimental design and primary analyses were preregistered at \url{https://aspredicted.org/6PX_5TD}.

\subsection*{Player recruitment}

The human players in our study were recruited from the graduate and undergraduate student body of Rensselaer Polytechnic Institute across various departments (see SM Section~3 for a demographic breakdown). Recruitment efforts consisted of emails sent to students from department heads, supplemented by wanted ads hung on campus. All recruitment material directed students to join a Discord channel dedicated to this study. Upon joining the channel, they were provided the terms and conditions of the study, the game rules, and instructions on how to join a game. 
SM Section 14 details participant preparation for the game, their essential compensations, and additional awards.

The game is hosted on customized playable software on mobile and PC browsers (available at \url{https://sociopinion.fly.dev}). See SM Section~16 for screenshots of the study's Discord channel and the game website, respectively.

\subsection*{Exit survey design}

Immediately after the 60-minute timer expires in the game, the human players are informed if they are one of the two winners, and then all are provided with a link to the exit survey. There are four different variants of exit surveys, two for each game type involving human players (\textit{HH} and \textit{AH}), with both having their own two variations: one for the winning players and one for the remaining players. There is no difference in the questions asked all survey variations. Still, slight differences in the auxiliary text shown on the surveys depend on the variation received.

All four versions include a message shown to players upon submitting the survey that informs them that this study has two human-involved game types: one with agents and one without. The survey specifies whether the player was playing with agents or just humans, depending on the type of game. Additionally, players who win the game are congratulated for winning the game within the survey.

The survey is divided into influence nominations, demographics, and payment information. For influence nominations, players are asked to subjectively identify which fellow players in their game were the most and least convincing. For this section, the players receive a list of usernames of the players involved, from which they can select one. The demographics are optional, and information about players' age, gender, and ethnicity is collected. Payment information was required.

\subsection*{Keyword analysis}

We identified 102 on-topic keywords relating to the game prompt regarding diet, diet nutritiousness, and climate consciousness. Additionally, keywords relating to opinion consensus dynamics, such as majority or minority opinions, were added. An initial selection of $36$ keywords was identified using the YAKE tool~\cite{campos2020yake}. This was done by deploying the tool on a pool of all $15,547$ messages sent by humans and agents. To encapsulate all aspects of the prompt and game dynamics in our dictionary of on-topic keywords, $66$ additional keywords were manually identified and added by authors. See SM Section~6 for a list of all the on-topic keywords used in this research.

\subsection*{Agent design}

The critical components of this study are the agents, as all analysis hinges on their ability to accurately represent the typical capabilities of the current-day LLMs we have chosen. We used a fusion of OpenAI's ChatGPT and GPT-4~\cite{achiam2023gpt} and Meta's Llama 2~\cite{touvron2023llama} to power the agents. GPT-4 and Llama 2's APIs are called for all conversational tasks, with ChatGPT's API being used to process auxiliary requests.

Our game follows the long tradition of simulating players in games with social interactions as agents, often endowed with personalities (\cite{10.1145/3586183.3606763} \cite{10.1145/3570945.3607359}\cite{https://doi.org/10.1111/infa.12472}).
Upon the initialization of a game, agents are instantiated with unique personas. Each persona consists of four traits. Each trait is represented by a digit $d_i$. Thus, in our games, there are $p=4$   digits. Each digit has a single value in the trait from the range $[0,m_i-1]$. $d_1$ represents the level of stubbornness with three values (stubborn, regular, and suggestible) determining how strongly agents defend their favorite diet in conversations, hence $m_1=3$. $d_2$ stands for grammar sophistication levels with three values (lowercase, perfect, and reduced punctuation), so $m_2=3$. The third digit represents initial personal confidence levels with four values (not very confident, somewhat confident, quite confident, and very confident), which indicate the strength of belief of the agent in its own diet choice, so $m_3=4$. The fourth digit represents an initial favorite diet with four values, hence $m_4=4$. The number of unique codes is $\prod^p_{i=1} m_i$.

Having four digits of personality $(d_1,d_2,d_3,d_4)$, we can compute the decimal value of the code for personality as follows: let $M_1=1$ and for $1<i<p, M_i=M_{i-1}*m_{i-1}$. Then, the decimal value of this code is $\sum_{i=1}^p d_i*M_i$. In our case $p=4$, and $M_2=3, M_3=9, M_4=36$. Therefore the code $(0,0,0,0)$ has decimal value of $0$ and the largest $(2,2,3,3)$ has decimal value of $2*M_1+2*M_2+3*M_3+3*M_4=2+6+27+108=143$, so there are $144$ unique personalities each represented by the decimal value ranging from $0$ to $143$.  

Often, not all trait values are compatible with each other in a persona. In our game, agents with the lowest confidence levels are unlikely to be stubborn, and those with the highest stubbornness are rarely suggestible. We do not allow such traits to be selected to accommodate such details. This eliminates 24 persona options. To preserve the uniqueness of personas and compatibility of the values of their traits, we create a bit vector $E$, of size $N=\prod^p_{i=1} m_i$, which is equal to the number of unique codes. For each  code $c$, we compute its decimal value denoted $d_c$ and initially set entries in bit vector $E[d_c]$ to either to $0$ if code $c$ has all digits compatible with each other and yet is unused a persona, and to $1$ otherwise. Then we randomly choose a persona with traits $c$, and if $E[d_c]=1$, then we repeat the creation of codes until the decimal code $c$ has $E[d_c]=0$, i.e., if code $c$ is unused. We assign code $c$ to the agent and increase $E[d_c]$ to one.

See SM Section~1 (including SM Tables~S1 and S2) for details on stubbornness levels and grammar sophistication levels, including our finding that the 'persuasiveness' of agents varied based on their grammar sophistication level and stubbornness level combination. See SM subsection~7.2 For full details on this persona assignment procedure. See SM subsections 7.3 - 7.5, showing that our agents had behavioral patterns as unique as human players (Including SM tables S9 and S10.)

We found that agent intelligence rapidly broke down when agents powered by a single LLM (i.e., GPT-4) for generating conversational responses were fed responses generated by other agents employing the same LLM in a shared game environment. This phenomenon is analogous to the "curse of recursion" shown in~\cite{shumailov2023curse}, where LLM-produced content was irreversibly decaying in quality as the amount of LLM-generated training data increased. To somewhat alleviate this issue and to improve response variety overall, we introduced agent functionality where a second LLM API (i.e., Llama 2) could be called for conversational tasks. Therefore, in situations where GPT -4's API is used, there is a $50\%$ chance of Llama 2's API being called to produce a response instead.

When a conversation is started between an agent and another player, the agent is given the following initial prompt:

\begin{displayquote}
"Let's play a game. You are in a debate contest with six participants, including yourself and me. Currently, you and I are in a one-on-one conversation. Four diet opinions are being considered: vegan, vegetarian, omnivorous and pescatarian. You believe the $m$ diet is the best compromise between nutritiousness and climate consciousness. You know that $n$ of the six participants, including yourself, share your opinion, and $p$ of six participants share my opinion. Your goal is to determine my opinion and try to convince me to change my opinion to $m$ or to ensure you are in a majority group that shares the same opinion. Keep your messages short and use informal and casual language. The goal of the debate is for both you and me to agree on the same diet instead of finding a middle ground or compromise. Do not repeat the same phrasing across your responses; aim for originality. Vary your sentence structure every time you respond."
\end{displayquote}

In the prompt, $m$ denotes the diet the agent currently favors. The agent is aware that there are $n$ players who share their opinion on diets, and there are $p$ players who share the views of the players with whom the agent is conversing. As the game progresses, the agent will attempt to keep track of which players had what opinion last, adjusting $n$ and $p$ accordingly. The variables $n$ and $p$ mimic a human's internal estimation of who is with or against their opinion in a game. Furthermore, like humans, agents are given agency in deciding who to share their estimations with during conversation. Subsequently, agents may collude with or block other players according to their knowledge of opinion distribution in-game, analogous to human memory in this context.

Agents were allowed to accept or reject an invitation to conversations and send their invitation to other players. Upon the start of a conversation, we found that early versions of the agents would out themselves as AI by being too verbose, formal, or naive. Therefore, we programmed the agents to open conversations with pre-written greetings and exchange diet opinions before presenting arguments. Agents have a time limit for a conversation. Thus, when this time limit expires, or the other player starts attempting to end the conversation, the agent sends meaningful farewell messages. In \textit{ah} conversation, the agents were also programmed to generate messages with withholding periods like those observed for human online speech. Additionally, agents were programmed to break up their responses into multiple sentences whenever grammatically possible, with each sentence being sent separately with a short delay to emulate the time it would take a human to type. We note that during \textit{aa} conversations, ancillary modifiers to text, like sentence splitting, are disabled (evident in the examples in SM Table S24).

With every new message posted in a conversation involving an agent, the agent will analyze the entire conversation and assess if the conversation has concluded. If it has, the agent making the assessment will trigger a farewell protocol and end the conversation. However, we found in early testing that agents were unreliable in determining when a conversation ended naturally. In the case of \textit{aa} conversations, this could lead to conversations that would last over the whole game time of 60 minutes. This, coupled with the drastic quality drop in message content for \textit{aa} conversations, prompted us to introduce a message limitation. Subsequently, we designed agents to have a randomly assigned "message budget" for the number of messages they can send and receive. Once the message budget is exceeded, the agent will be prompted to terminate the conversation regardless of the conversation state. For \textit{aa} conversations, both agents are assigned a budget between 12 and 16 messages. We also introduced this message limitation for agents in \textit{ah} conversations to impede humans from trapping agents in a conversation. However, we did not want agents to inadvertently prevent more extended conversations from developing fruitfully. Hence, the agent's budget for this conversation type is between 30 and 50 messages. See SM Section~8 for more details on the message budget and its effects on conversation length.

Once the farewell protocol had been triggered or the message budget was exhausted, agents were given this prompt:

\begin{displayquote}
"Tell me that you have decided to end the conversation. Be creative with your goodbye, using our conversation above as context. Do not say we will talk again. Do not confirm you have received these instructions."
\end{displayquote}

After the conversation has terminated, the entirety of the discussion is compiled and fed back to the agents involved. With this compilation provided as a prompt, the agents are requested to assess the confidence of both players, producing levels of perceived and personal confidence. The agents are also asked to determine if, based on the conversation, they should change their opinion, and if so, to what. When the agent engages in a conversation with a player with whom they conversed before, a history of the past discussion is fed into a ChatGPT API call to summarize. This Summary is added to the information initially given to the agent at the beginning of the conversation to allow the agent to simulate "memory" of the past discussion.

\subsection*{Agent detection categories}

We say that an ADI was an AI system provoked when a mechanistic weakness associated with the agents led to unprompted behavior, causing a detection. This includes cases where the agent began generating a self-introduction in the middle of a conversation, repeated something it had said earlier, and other similar failures. We say an ADI was AI language provoked when the agent's messages were perceived as too formal, lengthy, non-humanlike, and information-laden. On the other hand, some ADIs were not caused by the behavior of agents in conversation but by accusations from humans who had previously spoken with agents or were informed about them by a different human. Humans often began conversations by asking the other player if they were an AI. We call such ADIs humans "hunting." Finally, we consider \textit{hh} conversations in which agent activity was discussed as human-human spread ADIs.

\subsection*{Data Availability}
The data generated in this manuscript is available from the corresponding authors upon reasonable request.

\subsection*{Code Availability}
The code for the custom game website and AI used for the current study can be found at \url{https://github.com/Aganonce/LumityAI}.

\nocite{faul2009statistical, szudzik2006elegant, vehtari2023loo,  burkner2017brms}
\bibliography{bibliography}
\bibliographystyle{naturemag}

\subsection*{Author Contributions}
BKS and JF defined the focus of the paper and prepared and got approved for the IRB waiver; JF and MSM conceived the consensus game for the study and implemented and ran an interface to enable agents to play the game; BKS selected a team of authors and provided resources; BC determined the number of games to allow a statistically significant analysis to be performed, designed regression models, and ran these models, collected the results, and assessed statistical independence of results; BKS, JF, MSM conceived metrics and tests for predicting types of players from metrics; CM, JF, and MSM implemented and ran software that executed games, and collected data and computed metrics and tests; BC, BKS, JF, and MSM wrote the paper; all authors analyzed the data, interpreted the results, and edited and approved the final version of the paper.

\subsection*{Author Declaration}
All authors declare no competing interests.

\subsection*{Acknowledgments}
BC, JF, and BKS disclose support for the research of this work from the DARPA INCAS Program grant number HR001121C0165 and the NSF grant number BSE-2214216.

\end{document}


\baselineskip24pt

\maketitle

\textbf{This PDF file includes:}

\indent \indent
Supplementary Text

\indent \indent
Supplementary Figures 1 to 9

\indent \indent
Supplementary Tables 1 to 24

\newpage
\onehalfspacing 


\section{Agent configuration}

The functional capability of our conversational agents is an essential aspect of our experiment. We leveraged two LLM models, Open AI's GPT-4 and Meta AI's Llama 2, to generate agent responses. Notably, we curated the protocols and personalities of our agents to ensure every agent in a game had a meaningfully different personality. Specifically, we created nine different configurations, and each agent was randomly assigned to one at the beginning of the game. Below, we describe the agent stubbornness levels and grammar sophistication levels we designed. We also measure the persuasiveness of the nine configurations.

\subsection{Agent stubbornness and grammar sophistication levels}
We designed three stubbornness levels for the agents: suggestible, regular, and stubborn. All three levels had a standard directive to bring conversations to a mutual agreement on one of the four diet opinions instead of a nuanced ``common ground''. This discouraged the agents from agreeing with the other participants and pushed them to debate rather than converse. 

For the regular level of stubbornness, the prompt provided to the agent was, "Remember that the goal of the debate is for both you and me to agree on the same diet instead of finding a middle ground or compromise." For the stubborn level of stubbornness, the prompt was, "IMPORTANT: Your personality for this conversation is stubborn and unyielding. You must write strong rebuttals against my arguments to convince me to change my opinion at all costs. Remember that the goal of the debate is for both you and me to agree on the same diet instead of finding a middle ground or compromise. However, be inviting if I want to change my opinion to support [your] diet." The prompt for the suggestible level of stubbornness was, "IMPORTANT: Your personality for this conversation is suggestible, and you can change your opinion if I make logical arguments against it. Furthermore, you are willing to change your opinion if that decision makes you part of the majority in the game", followed by the regular level of stubbornness text. These stubbornness level prompts worked in conjunction with several other contextualizing prompts.

When presented with compelling arguments, the suggestible level was designed to change their opinion in favor of the other participant. The stubborn level was intended to present solid arguments and rebuttals to shift the other participant's opinion. In the experiment, about half the agents across all games were regular level, while one-quarter of agents were suggestible and one-quarter were stubborn.

\newcolumntype{g}{>{\columncolor{lightgray}}c}
\begin{table*}[h]
    \centering
    \renewcommand{\arraystretch}{1.25}
    \begin{tabular}{|g|*{3}{c|}}
        \hline
        \rowcolor{lightgray}
        \textit{Num. of agents} & \textbf{Suggestible} & \textbf{Regular} & \textbf{Stubborn}  \\
        \hline
        \textbf{Lowercase} & 4 & 11 & 3  \\
        \hline
        \textbf{Regular} & 4 & 3 & 8 \\
        \hline
        \textbf{Reduced Punc.} & 6 & 8 & 4 \\
        \hline
    \end{tabular}
    \caption{\textbf{Agent configuration breakdown.} The number of agents in each of the nine agent configuration groups is based on stubbornness-grammar pairings. Every agent from \textit{AH} games is included. Three agents were in each of the 17 \textit{AH} games.}
    \label{tab: configurations} 
\end{table*}

We also had three grammar sophistication levels: lowercase, perfect, and reduced punctuation. The lowercase agents use lowercase type only and avoid punctuation at the end of sentences to mimic the vocabulary of modern instant messaging. The perfect agents make no changes to the default agent punctuation, which is usually grammatically correct. The reduced punctuation agents maintain the correct type casing but remove most punctuation characters. 

We define an agent configuration as a stubbornness level and grammar sophistication level pair. As shown in Table \ref{tab: configurations}, there are nine possible agent configurations.

\subsection{Agent persuasiveness}
We develop \textbf{``Persuasiveness''} as an effectiveness metric for agents in \textit{AH} games computed on a per-conversation basis. It is based on two rules. First, if an agent convinced a participant to change their opinion to match their own after a conversation, the agent gains 3 points. Second, if there was no opinion change, the agent gains points equal to the reduction in the other participant's confidence score (since confidence is between 1 and 4, it can be most reduced by 3). Thus, the persuasiveness score reflects the agent’s ability to influence others. An agent can get a maximum score of 3 and a minimum score of 0 per conversation. A perfectly persuasive agent would achieve an average score of 3 points across all its discussions.

\begin{table*}[h]
    \centering
    \renewcommand{\arraystretch}{1.25}
    \begin{tabular}{|g|*{3}{c|}}
        \hline
        \rowcolor{lightgray}
        \textit{\% Persuasive} & \textbf{Suggestible} & \textbf{Regular} & \textbf{Stubborn}  \\
        \hline
        \textbf{Lowercase} & 23.83\% & \textbf{64.64\%} & 33.33\%  \\
        \hline
        \textbf{Regular} & 33.33\% & 11\% & 41.67\% \\
        \hline
        \textbf{Reduced Punc.} & \textbf{61.1\%} & \textbf{70.73\%} & 0\% \\
        \hline
    \end{tabular}
    \caption{\textbf{Persuasiveness scores between 0 and 100 percent for each agent configuration.} Scores indicate the percentage shift in opinion confidence or changing of opinions as described in the text. The top three configurations are in bold.}
    \label{tab: persuasiveness} 
\end{table*}

We computed the persuasiveness score for each of the 51 agents across all \textit{AH} games. We aggregated the scores for each agent configuration and converted them to percentages, ranking each configuration somewhere between \(0\%\) to \(100\%\) persuasive. Table \ref{tab: persuasiveness} shows the outcomes. Three agent configurations stand out as most compelling, the best one being the Regular-Reduced punctuation configuration, which was \(70.73\%\) persuasive (meaning these agents affected an average shift of 2.125 confidence points toward their opinion). The reduced punctuation grammar sophistication levels and regular personalities were the most convincing.

An important caveat is that though these scores show the most persuasiveness configurations in our experiment, they may not indicate future performance. Participants' demographics, random luck in initial opinion assignment, familiarity with AI language models, the outcomes of a conversation between two agents, and many other confounding qualitative factors may affect how effective each configuration can be. Also, as described in the above sections, granted them a base level of effectiveness.

\section{Holding periods and response times}

As participants in the study participate in one-on-one conversations, there are two temporal metrics through which the activity of each participant in our research can be measured. The first is \textbf{``holding period''}, which we define as the amount of time spent by a participant to send a block of contiguous messages (which we call a message chain) without interruption from the other participant. This metric measures the time spent by a participant for writing a reply that involves more than just a single message. Message chains are a common way for human participants to communicate on instant messenger applications, and this behavior is found on our platform, too. This is because our platform looks and feels just like a traditional instant messenger or SMS app when conversing. 

The second metric to measure is \textbf{``response time''}, which is the time between two messages from different participants. In other words, it is the time taken for a participant to send a reply after receiving a message. Response time for a reply is measured from the timestamp of the last message from the previous participant to the timestamp of the first response of the current participant. 

Response time and holding period are essential to get a fuller picture of the amount of activity in a conversation. These metrics provide a complete picture of the conversation's activity as a function of time. A caveat is that holding periods are essentially static for the agents, who are programmed always to send a message chain of 2-4 messages separated by 3 seconds each. Response times for agents are also not of interest, as the time taken for an agent to generate a response will also be static. This caveat and the fact that the agents are not programmed to be sensitive to human response times or holding periods, except a reminder sent if a conversation has been inactive, means that analyzing these metrics for \textit{aa} conversations is irrelevant. 

We start by measuring holding periods and response times for every conversation across every game in our study. After that, we measure the statistical properties of this data, such as averages and medians. We analyze three conversation types: \textbf{1. \textit{hh} conversation from \textit{HH} games}, \textbf{2. \textit{hh} conversations from \textit{AH} games} where the latent effect of agent presence may have affected human conversation trends, and \textbf{3. \textit{ah} conversations from \textit{AH} games}. We visualize their properties and compare each configuration pair for the significance of differences in holding period and response time trends.

\subsection{Statistical properties of holding periods and response times}
\begin{figure}[t]
    \centering
    \includegraphics[width=6in]{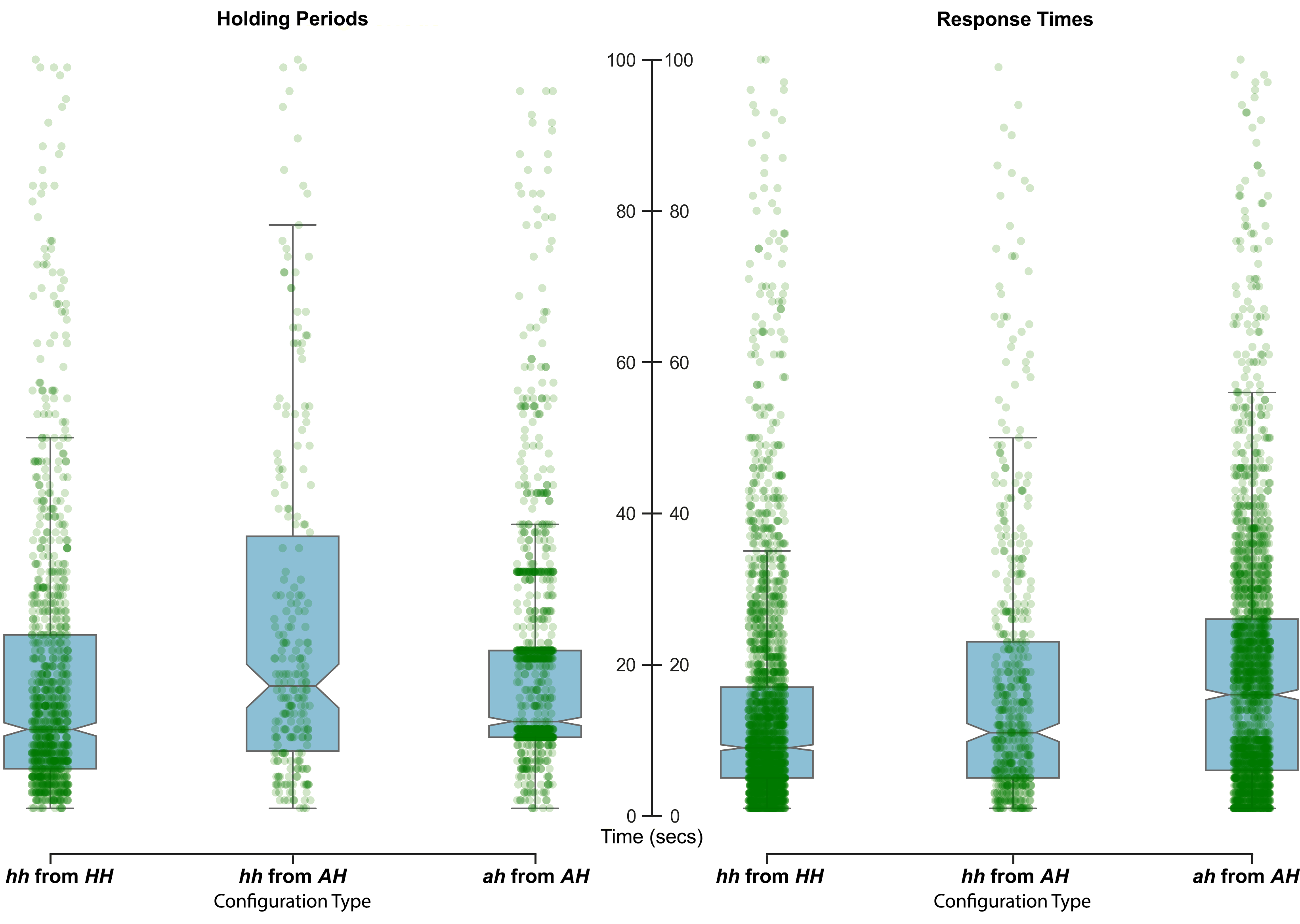}
    \caption{\textbf{Boxplots for Holding Periods and Response Times.} We analyze three conversation types: \textit{hh} conversations from \textit{HH} games, \textit{hh} conversations from \textit{AH} games and \textit{ah} conversations from the same games. Each green dot indicates a message instance's holding period or response time.}
    \label{fig: S1}
\end{figure}

We tackle each of our three chosen conversation types individually using the following process: We divide the messages sent in all conversations of a kind into message chains and single messages. We measure the holding period of every message chain and then statistically characterize that data. We then measure the response times and characterize them, too. We use Figure \ref{fig: S1} to visualize the interquartile ranges as boxplots. Each box captures 50\% of the data in between the first (Q1) and third (Q3) quartile (i.e., in between the \(25^{th}\) and \(75^{th}\) percentiles), with the line inside the box indicating the median. The whiskers on either side indicate the minimum and maximum as per boxplot convention, the minimum being \(Q1 - 1.5 * IQR\) and maximum being \(Q3 + 1.5 * IQR\), with the dots outside whiskers indicating outliers that fall beyond the maximum.

\subsubsection{\textit{hh} conversations from \textit{HH} games}
For \textit{hh} conversations from \textit{HH} games, there were 2,516 message instances, of which 1,028 were message chains consisting of more than one message. The average holding period for these chains was 21.99 secs, with the median being 11.50 secs and the mode being 7 secs. The holding period boxplot for HO in Figure \ref{fig: S1} shows that about 50\% of the chains had a holding period between 6 and 24.25 secs, with the majority of them not exceeding 51 secs as demonstrated by the extent of the top whisker. There were 91 outlier instances where the holding period exceeded 51 secs.

For \textit{hh} response times, 2,305 message instances had a response time greater than zero. The mean response time was 15.04 secs, with the median being 9 secs and the mode being 5 secs. The response time boxplot for \textit{hh} conversations shows that about half of all responses had response times between 5 and 18 secs, with most not exceeding the top whisker of 37 secs. There were 192 outliers where the response times exceeded the top whisker. A minuscule amount of outliers exceeded 500 secs and were discarded from analysis as erroneous.

\subsubsection{\textit{hh} conversations from \textit{AH} games}
For \textit{hh} conversations from \textit{AH} games, there were 607 message instances, with 250 being message chains. The average holding period was 32.12 secs, with the median being 17.5 secs and mode being 4 secs. The holding period boxplot in Figure \ref{fig: S1} shows that half the chains had a holding period between 10 and 40.5 secs, most of which were under the top whisker at 86 secs. There were 17 outlier instances.

Regarding response times, 564 message instances had a valid response time (greater than zero). The mean response time was 21.2 secs, with the median being 11 secs and the mode being 5 secs. The \textit{hh} from \textit{AH} response time boxplot shows that half the time, people took between 5 and 25.25 seconds to respond, with most not exceeding 55 seconds. There were 49 outliers.

\subsubsection{\textit{ah} conversations from \textit{AH} games}
For \textit{ah} conversations from \textit{AH} games, there were 2,438 message instances, of which 1,078 were message chains. These chains show an average holding period of 21.75 secs, a median of 14 secs, and a mode of 10 secs. The \textit{ah} from \textit{AH} holding period boxplot in Figure \ref{fig: S1} shows that about half of all chains had a holding period between 10 and 13.75 secs, with an IQR-based maximum of 44 secs. There were 89 outlier instances where the holding period exceeded the maximum.

For response times, 2,237 message instances had a response time greater than zero. The mean response time was 22.18 secs, with the median being 16 secs and the mode being 1 sec. The response time boxplot shows that approximately 50\% of the responses had response times between 6 and 27 secs, with most not exceeding 58 secs. There were 127 outliers. 

\subsection{Comparison of conversation type pairs}

\begin{table*}[t]
    \centering
    \renewcommand{\arraystretch}{1.25}

    \begin{tabular}{|*{4}{c|}}
        \cline{2-4}
        \multicolumn{1}{c|}{} & \textbf{Pair 1} & \textbf{Pair 2} & \textbf{Pair 3}  \\
        \hline
        \textbf{Hyp. 1: Holding Periods} & \(\mathbf{1.819e^{-05}}\) & \(0.941\) &  \(\mathbf{1.189e^{-06}}\)  \\
        \hline
        \textbf{Hyp. 2: Response Times} & \(\mathbf{1.014e^{-09}}\) & \(\mathbf{1.940e^{-20}}\) & \(0.493\) \\
        \hline
    \end{tabular}
    \caption{\textbf{HP-RT hypothesis testing.} Independent two-sample t-test outcomes for each pair of conversation types for the two hypotheses. Bold values indicate a statistically significant value difference between the indicated pair of conversation types. Values not in bold indicate that the null hypothesis was accepted. Pair 1 is \textit{hh} from \textit{HH} and \textit{AH}, Pair 2 is \textit{hh} from \textit{HH} and \textit{ah} from \textit{AH}, Pair 3 is \textit{hh} and \textit{ah} from \textit{AH}.}
    \label{tab:tests_hprt} 
\end{table*}

We take the entire population of holding periods and response time data for each possible pair of conversation types and perform independent two-sample T-tests to find if the types are significantly different. We calculate population variance and perform Welch's t-test if variances are unequal. The null hypotheses we propose are:

\textbf{Holding period null hypothesis:} There is no statistically significant difference in the holding periods of message chains between conversations from conversation type X and conversation type Y.

\textbf{Response time null hypothesis:} There is no statistically significant difference in the response times of messages between conversations from conversation types X and Y.

We test three pairs of conversation types with these hypotheses. Pair 1 is \textit{hh} from \textit{HH} and \textit{AH}, Pair 2 is \textit{hh} from \textit{HH} and \textit{ah} from \textit{AH}, Pair 3 is \textit{hh} and \textit{ah} from \textit{AH}. The results of the t-tests are shown in Table \ref{tab:tests_hprt}. From the outcomes, there was a significant difference in holding periods and response times for Pair 1, showing that the latent effect of agent presence affected \textit{hh} conversations in \textit{AH} games, making them meaningfully different compared to \textit{HH} games with no agents present in the environment. We see a significant difference in holding periods for Pair 3 and response times for Pair 2, which shows that agents in conversations with humans facilitated new dynamics that greatly affected human behavioral trends. Our keyword and productivity analyses expose some of these new dynamics. 

\section{Demographic analysis}

\begin{figure}
     \centering
     \begin{subfigure}[b]{0.45\textwidth}
         \centering
         \includegraphics[width=\textwidth]{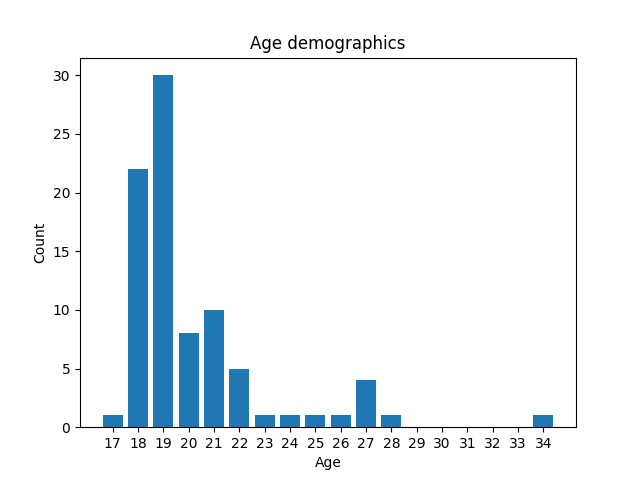}
         \caption{}
         \label{fig:demographics_a}
     \end{subfigure}
     \hfill
     \begin{subfigure}[b]{0.45\textwidth}
         \centering
         \includegraphics[width=\textwidth]{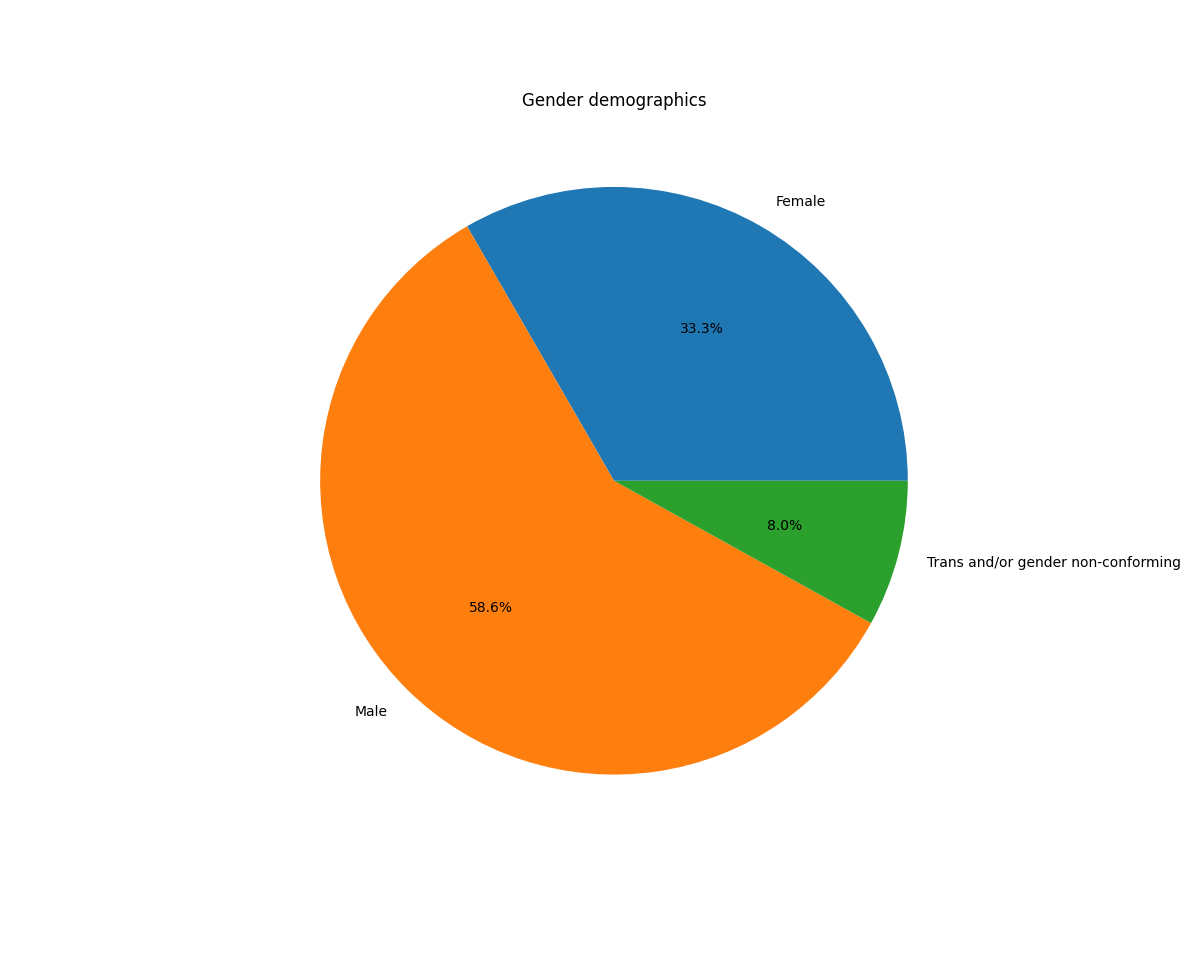}
         \caption{}
         \label{fig:demographics_b}
     \end{subfigure}
     \hfill
     \begin{subfigure}[b]{0.45\textwidth}
         \centering
         \includegraphics[width=\textwidth]{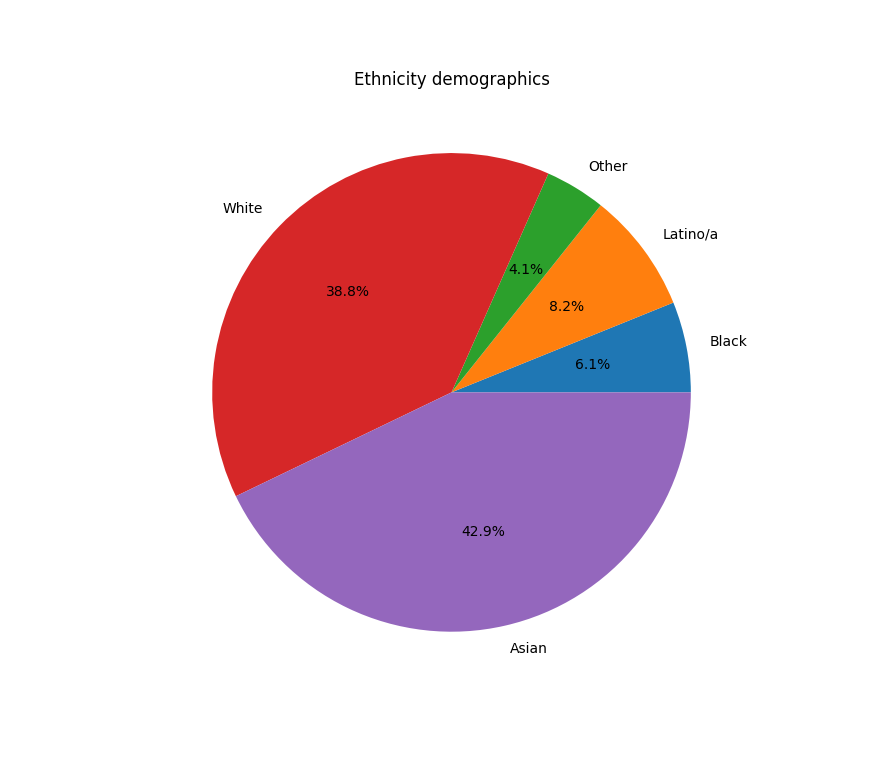}
         \caption{}
         \label{fig:demographics_c}
     \end{subfigure}
        \caption{\textbf{Demographic breakdown of participants.} (\textbf{\subref{fig:demographics_a}}) Histogram of participant ages. (\textbf{\subref{fig:demographics_b}}) Pie chart of the genders chosen. (\textbf{\subref{fig:demographics_c}}) Pie chart of the ethnicities chosen.}
        \label{fig:demographics}
\end{figure}

Demographics on age, gender, and ethnicity were voluntarily collected on participants who completed a game and filled out the exit survey where the optional demographic questions were presented. Within the exit survey, participants were asked to input a number for their age and select one option from a dropdown for gender. For ethnicity, they were also given multiple options, with the ability to select more than one. The options for gender were ``male'', ``female'', ``Trans and gender non-conforming'', and ``None of the above''. For ethnicity, the options were ``Latino/a'', ``White'', ``Black'', ``Asian'', ``Native American/Alaska Native'', ``Native Hawaiian/Other Pacific Islander'', and ``Other''. Overall, 91 participants completed the demographic portion of the exit survey. Their results are summarized in Figure~\ref{fig:demographics}. For Figure~\ref{fig:demographics}\subref{fig:demographics_a}, the average age was $23.38$ with a standard deviation of $4.516$. According to Figure~\ref{fig:demographics}\subref{fig:demographics_b}, the primary gender identity of the participants was male, with the majority of participants also claiming Asian ethnicity (Figure~\ref{fig:demographics}\subref{fig:demographics_c}).

\section{Human behavior before and after an agent detection incidents}

We want to analyze human behavior before and after an agent detection incident, as established in the main manuscript. Specifically, we compare human behavior related to our study's primary metrics: opinion change, perceived confidence, and personal confidence. To perform this analysis, we collected all conversations that involved participants who either triggered an agent detection incident or were in a conversation where an incident was triggered. With this smaller data set, we label each discussion with a 0 or 1, denoting that it occurred before (0) or after (1) the agent discovery event. We then reran our opinion-changing, perceived confidence, and personal confidence regression models using this subset of data, using this dichotomous ``detected'' variable as the fixed effect. We model random effects in the same way in these models as we do in the main manuscript versions. We list the results of the opinion change, perceived confidence, and personal confidence models in the following tables, respectively: Table~\ref{tab:ad_opinion_change_model}, Table~\ref{sm_tab:ad_perceived_model}, and Table~\ref{sm_tab:ad_change_personal_model}.

\begin{table}[htp!]
\centering
\begin{adjustbox}{width=.7\columnwidth,center}
\begin{tabular}{|rrr|}
  \hline
  Predictors & Odds ratios & CI (95\%)  \\
  \hline
  intercept & 0.07 & 0.00 - 0.36 \\
  Detected & 0.29 & 0.03 - 1.99 \\
   \hline
\end{tabular}
\begin{tabular}{|l|l|}
     \hline
    	Observations & 91 \\
        $N_{game}$ & 11 \\
        $N_{participant}$ & 18 \\
        $\sigma^2$ & 3.29 \\
        $\tau_{00\ game}$ & 1.74 \\
        $\tau_{00\ participant}$ & 1.18 \\
        Marginal $R^2$ & 0.006 \\
        Conditional $R^2$ & 0.122 \\
	\hline
	\end{tabular}
 \end{adjustbox}
 \caption{\textbf{Opinion changing before and after agent detection.}
This table defines the parameters and coefficients of this Bayesian multilevel model, presented as odds ratios. The detected coefficient shows how much more or less opinion switching occurs after the participant detects agents in the study.}

 \label{tab:ad_opinion_change_model}
\end{table}

\begin{table}[htp!]
\centering
\begin{adjustbox}{width=.9\columnwidth,center}
\begin{tabular}{|rrr|}
  \hline
  Predictors & Odds Ratios & CI (95\%)  \\
  \hline
  Perceived confidence 0|1 threshold & 0.04 & 0.01 - 0.18 \\
  Perceived confidence 1|2 threshold & 0.18 & 0.05 - 0.69 \\
  Perceived confidence 2|3 threshold & 0.49 & 0.14 - 1.82 \\
  Perceived confidence 3|4 threshold & 1.63 & 0.49 - 6.24 \\
  Detected & 0.59 & 0.21 - 1.64 \\
   \hline
\end{tabular}
\begin{tabular}{|l|l|}
     \hline
        Observations & 91 \\
        $N_{game}$ & 11 \\
        $N_{particpiant}$ & 18 \\
        $\sigma^2$ & 0.25 \\
        $\tau_{00}$ & 1.47 \\
        ICC & 0.15 \\
        Marginal $R^2$ & 0.017 \\
        Conditional $R^2$ & 0.262 \\
	\hline
	\end{tabular}
 \end{adjustbox}
 \caption{\textbf{Perceived confidence before and after agent detection.}
This table defines the parameters and coefficients of this Bayesian multi-level model, with the coefficients presented as odds ratios.}
\label{sm_tab:ad_perceived_model}
\end{table}

\begin{table}[htp!]
\centering
\begin{adjustbox}{width=.9\columnwidth,center}
\begin{tabular}{|rrr|}
  \hline
  Predictors & Odds Ratios & CI (95\%)  \\
  \hline
  Personal confidence change -1 | 0 threshold & 0.06 & 0.02 - 0.17 \\
  Personal confidence change 0 | +1 threshold & 2.94 & 1.22 - 7.02 \\
  Detected & 0.46 & 0.16 - 1.24 \\
   \hline
\end{tabular}
\begin{tabular}{|l|l|}
     \hline
        Observations & 91 \\
        $N_{game}$ & 11 \\
        $N_{particpiant}$ & 18 \\
        $\sigma^2$ & 0.01 \\
        $\tau_{00}$ & 0.27 \\
        ICC & 0.03 \\
        Marginal $R^2$ & 0.027 \\
        Conditional $R^2$ & 0.057 \\
	\hline
	\end{tabular}
 \end{adjustbox}
 \caption{\textbf{Change in personal confidence before and after agent detection.}
This table defines the parameters and coefficients of this Bayesian multi-level model, with the coefficients presented as odds ratios.}
\label{sm_tab:ad_change_personal_model}
\end{table}

Each of these regressions shows a ``detected'' odds ratio estimate with wide confidence intervals that are at the same time significantly lower and higher than 1 (which would indicate no relationship). This, coupled with the considerably reduced dataset size of 91 observations (compared to 1284 in the entire dataset), suggests that we do not have enough data on these before and after agent detection conversations to estimate the effect this discovery had on participant behavior.

\section{Agent grammar-stubbornness effects on agent detection incidents}

In addition to our main manuscript's analysis on agent detection incidents (ADIs) and the reasons behind their occurrence, we also investigated if any of the nine agent grammar-stubbornness configurations (outlined in SM section 1) were more likely to cause ADIs than others. To do this, we looked at the detection ratio for each agent configuration across all \textit{AH} games with ADIs, shown in Table \ref{sm_tab:adi_configurations}. From the outcomes, we see that at least half of all agents in each agent configuration (except for the ``Regular-Regular'' configuration, which only had one agent) triggered detection. No one configuration was much weaker than the others, implying that agent stubbornness and grammar modalities did not affect detection significantly.

\begin{table*}[h]
    \centering
    \renewcommand{\arraystretch}{1.25}
    \begin{tabular}{|g|*{3}{c|}}
        \hline
        \rowcolor{lightgray}
        \textit{Num. of agents} & \textbf{Suggestible} & \textbf{Regular} & \textbf{Stubborn}  \\
        \hline
        \textbf{Lowercase} & 1 of 1 & 4 of 6 & 2 of 2 \\
        \hline
        \textbf{Regular} & 1 of 2 & 0 of 1 &2 of 4 \\
        \hline
        \textbf{Reduced Punc.} & 3 of 5 & 4 of 5 & 1 of 1 \\
        \hline
    \end{tabular}
    \caption{\textbf{Agent configuration to ADI ratios.} Each cell represents a configuration type that would be assigned to an agent. The value in each cell gives the ratio of the number of detected agents to the total number of agents with that configuration. Only agents from \textit{AH} games with ADI events were counted. Three agents were in each of the nine \textit{AH} games with ADIs; hence, 27 agents are considered here.}
    \label{sm_tab:adi_configurations} 
\end{table*}

\section{Keyword dictionary}

The $36$ keywords detected by the YAKE tool are:

\begin{displayquote}
farm, food, products, plants, pest, overfishing, foods, world, nutritional, supplements, animal, team, environmentally, climate, nutrition, plant, fishing, eat, planet, plant-based, diets, seafood, impact, farms, water, meats, sustainable, environmental, farming, veganism, meat, agriculture, reduce, eating, ocean, Omni
\end{displayquote}

\noindent
The $66$ supplementary keywords identified by manual inspection of the messages are:

\begin{displayquote}
vegetarian, nutritious, welfare, swaying, omnivorous, balance, vitamins, disease, sides, take, pescatarian, healthy, emissions, variety, protein, footprint, diet, more, case, smart, sea, vegetarianism, fish, ethics, se, omnivore, apple, bread, veggies, minority, chipotle, environment, better, carrots, vegan, page, gas, adopt, ethical, beef, health, fruits, peas, animals, nutritiousness, pasta, pescetarianism, convinced, convince, best, sustainability, majority, eggs, carbon, deforestation, nutrients, steak, roots, vegetables, milk, rice, carbohydrates, bagel, eco, sway, carbs
\end{displayquote}

\section{Minimum number of games and participants}

During the design phase of the study, an important consideration was the minimum number of games and participants per game type, which we would need to detect significant differences between behaviors in different game types and between participants within a single game type. To do this, we performed a priori power analysis using G*Power [15, 28]. We aimed to get $\sim0.8$ predictive power with our target analysis against participant and interaction level data. The main parameters for our power analysis were $\alpha$ (type 1 error rate) $ = 0.05, \beta $ (type 2 error rate) $ = 0.2,$ with a medium effect size.

\begin{figure}
     \centering
     \begin{subfigure}[b]{0.45\textwidth}
         \centering
         \includegraphics[width=\textwidth]{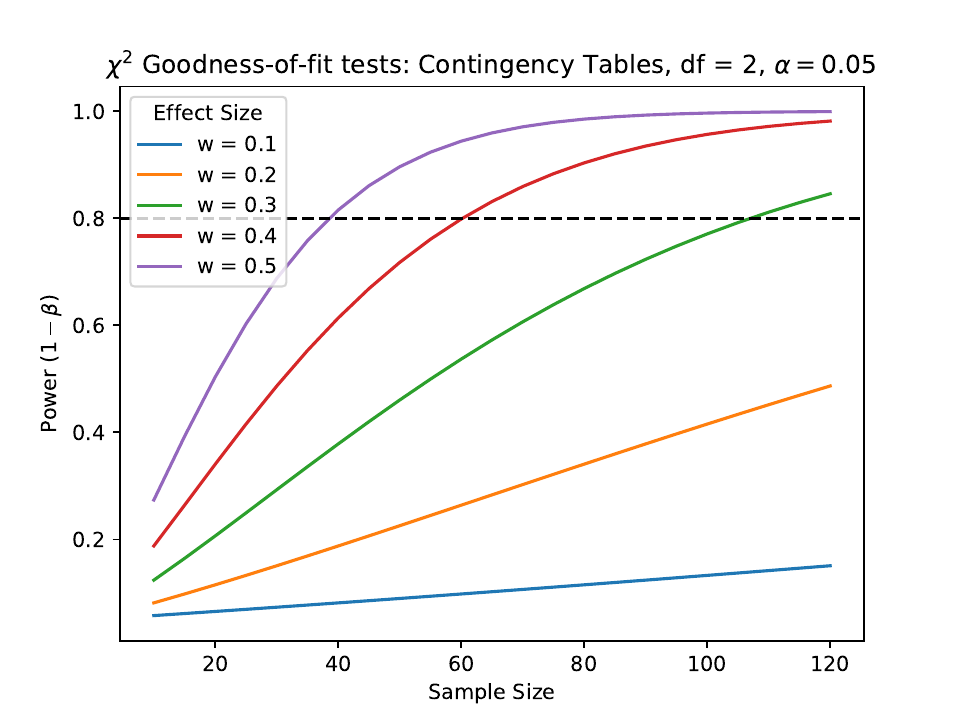}
         \caption{}
         \label{fig:power_analysis_a}
     \end{subfigure}
     \hfill
     \begin{subfigure}[b]{0.45\textwidth}
         \centering
         \includegraphics[width=\textwidth]{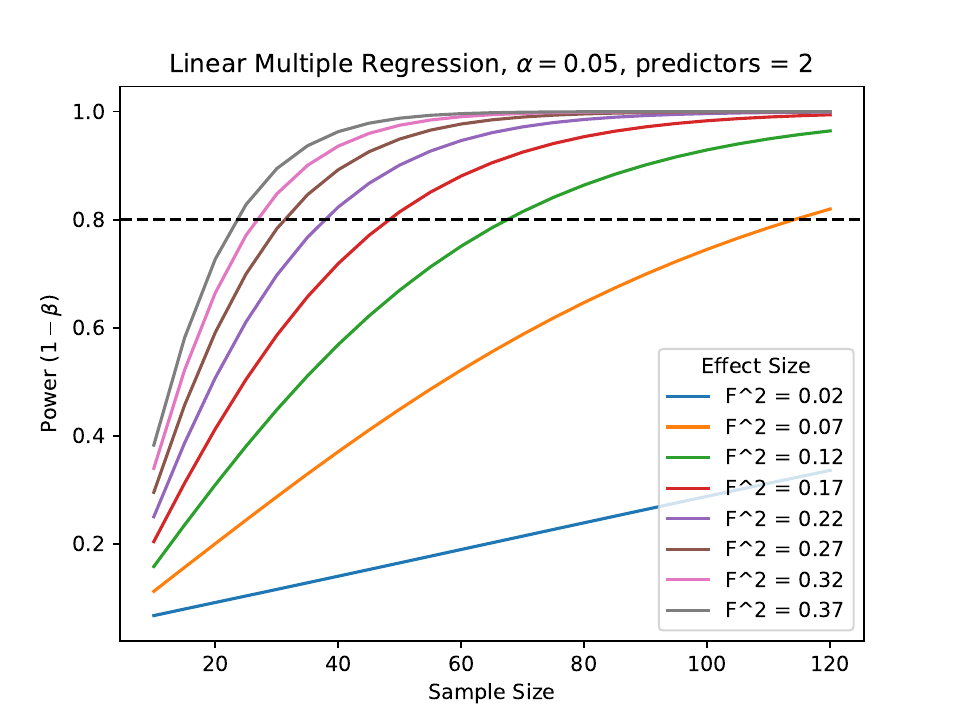}
         \caption{}
         \label{fig:power_analysis_b}
     \end{subfigure}
     \hfill
     \begin{subfigure}[b]{0.45\textwidth}
         \centering
         \includegraphics[width=\textwidth]{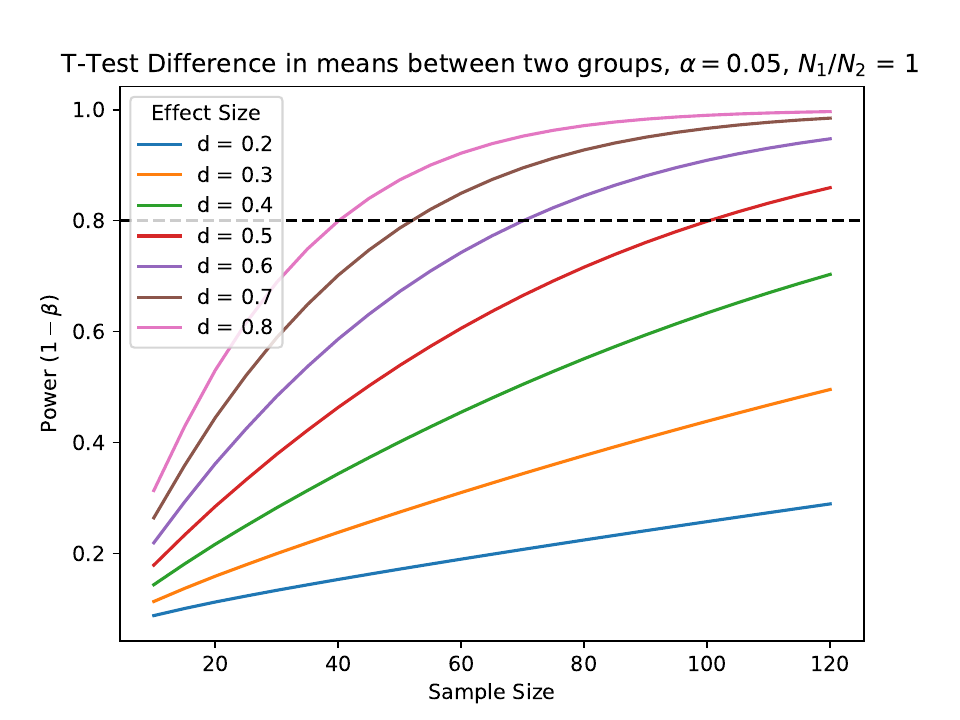}
         \caption{}
         \label{fig:power_analysis_c}
     \end{subfigure}
        \caption{\textbf{Power Analysis.} Each plot shows the projected power of different target analyses for different sample sizes/effect sizes. Effect sizes go from a small effect size for the given analysis to a large effect size. Our target is to have .8 power for a medium effect size for each analysis, which we can achieve with 120 samples (participants) per pairwise comparison of groups.}
        \label{fig:power_analysis}
\end{figure}

Our target analysis is described in our pre-registration. It varies depending on the level of study (game level, participant level, interaction level). Still, it can be summarized as a multi-level regression, a t-test comparison between groups, or a $\chi^2$ goodness-of-fit test. Figure \ref{fig:power_analysis} shows the power of this analysis over a range of sample sizes and effect sizes. To detect a medium effect size (0.5) with .8 power, we would require 120 combined samples from both groups. This analysis gave us a target number of games per game type at 10. At 6 participants per game, we would have 60 participants per game type. We gathered additional samples of agent-human interaction to enhance the predictive power of within-group analysis of \textit{AH} games, collecting 17 \textit{AH} game samples, over 100 participant data points. 

While we may have gathered extra participants for the \textit{AH} games, it is important to note that throughout our study, some human participants either failed to engage in discussion with other participants or disconnected from the study entirely. These participants lower the total number of human participants from 111 to 97. This amounts to 6 missing participants from agent-human games and eight from human-human games. This implies that eleven AH games had the full population of humans (three humans) while the remaining six had two humans instead of three after accounting for attrition. There was one game (Game ID 1225242) where a pair of humans engaged in a single conversation for the entire hour and did not record their opinions afterward. Their messages were analyzed but not included in other analyses, giving us 95 human participants for participant level and conversation level analysis. For HH, four HH games had six humans, four games had five humans, and two games had four humans after accounting for attrition.

While these missing data points lower the power of our analysis, resulting in a lower probability for us to be able to reject our null models in the case of a true signal, much of our analysis is done on the conversation level data. Each participant partakes in multiple conversations throughout a game, and therefore, each participant contributes multiple data points. This results in us having, in most cases, higher power on our target analysis.

\subsection{Analysis of Game Population Differences}

In HH and AH games, there were some instances where humans would leave the game before starting any conversations, which we call human attrition. For these cases, we analyze how our observations would change if games with human attrition were removed from this section's key analyses. The main manuscript and SM analyses, such as opinion change, confidence, and conversational productivity, are already based on just the present players, as the humans that left before playing did not generate any data and, therefore, did not affect those measures. We also carefully used the correct 'N' for these analyses. However, the accuracy of pairwise analyses needs to confirm that human attrition did not lead to statistically significant differences in those key characteristics when we compare agent and human behavior.

\begin{table}[t]
\begin{tabular}{|l|lllllll|llll|}
\hline
Participant 1 & \multicolumn{7}{c|}{Human}                                                                         & \multicolumn{4}{c|}{Agent}                                                                         \\ \hline
Participant 2 & \multicolumn{5}{c|}{Human}                                 & \multicolumn{2}{c|}{Agent}            & \multicolumn{2}{c|}{Human}                                 & \multicolumn{2}{c|}{Agent}            \\ \hline
Population    & \multicolumn{4}{l|}{All}    & \multicolumn{1}{l|}{Not All} & \multicolumn{1}{l|}{All}    & Not All & \multicolumn{1}{l|}{All}    & \multicolumn{1}{l|}{Not All} & \multicolumn{1}{l|}{All}    & Not All \\ \hline
Stay          & \multicolumn{4}{l|}{161}    & \multicolumn{1}{l|}{95}      & \multicolumn{1}{l|}{91}     & 39      & \multicolumn{1}{l|}{58}     & \multicolumn{1}{l|}{34}      & \multicolumn{1}{l|}{523}    & 45      \\ \cline{1-1}
Switch        & \multicolumn{4}{l|}{25}     & \multicolumn{1}{l|}{14}      & \multicolumn{1}{l|}{2}      & 2       & \multicolumn{1}{l|}{36}     & \multicolumn{1}{l|}{7}       & \multicolumn{1}{l|}{139}    & 13      \\ \cline{1-1}
Stay ratio    & \multicolumn{4}{l|}{0.8655} & \multicolumn{1}{l|}{0.8716}  & \multicolumn{1}{l|}{0.9785} & 0.9512  & \multicolumn{1}{l|}{0.6170} & \multicolumn{1}{l|}{0.8293}  & \multicolumn{1}{l|}{0.7900} & 0.7759  \\ \cline{1-1}
Switch ratio  & \multicolumn{4}{l|}{0.1344} & \multicolumn{1}{l|}{0.1284}  & \multicolumn{1}{l|}{0.0215} & 0.0488  & \multicolumn{1}{l|}{0.3830} & \multicolumn{1}{l|}{0.1707}  & \multicolumn{1}{l|}{0.2099} & 0.2241  \\ \hline
\end{tabular}
\caption{Opinion switching dynamics between full and partial population AH games}
\label{tab:opinion_change_attrition_comparison}
\end{table}

\begin{figure}[t]
    \centering
    \includegraphics[width=0.8\textwidth]{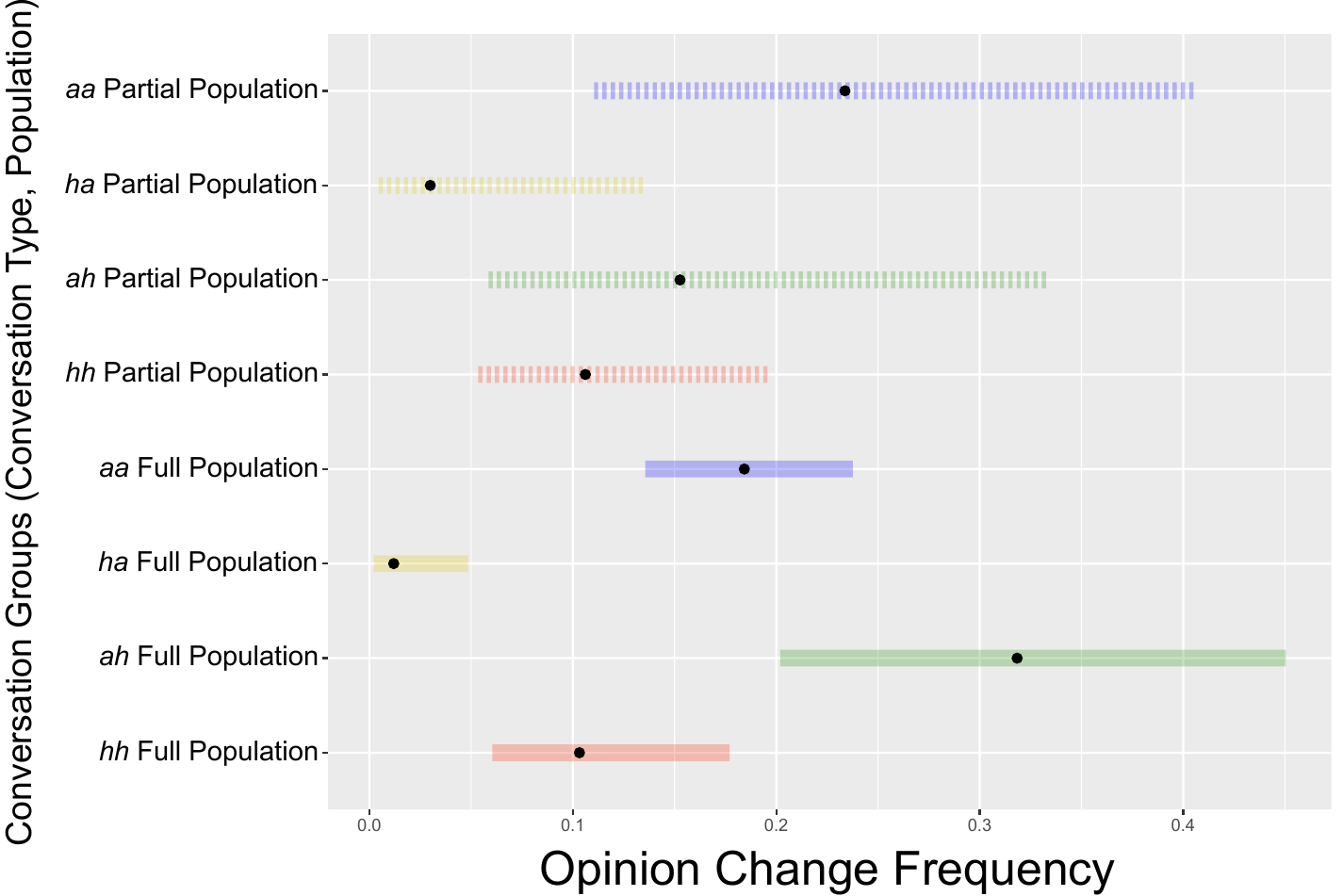}
    \caption{The opinion change trends between full and partial games. Here, we show the estimated opinion change frequency for both full and partial population games. The same conversation groups are shown in the same color, with partial games having a dashed line. We can see that all coefficients of the partial games are well within the CI of the full games, with most mean estimates being very similar. Note, \textit{hh}, \textit{ha}, \textit{ah}, and \textit{aa} all indicate conversations between two participants where the first letter is the participant type we are measuring opinion change on and the second is the participant type of the conversational partner.}
    \label{fig:opinion_change_partial_vs_full}
\end{figure}

Opinion change trends were similar between games with full and partial populations as captured in table \ref{tab:opinion_change_attrition_comparison}. There were no significant differences in opinion-switching behavior by the numbers. The overlapping confidence intervals of the full population conversations and their partial population counterparts in Fig. \ref{fig:opinion_change_partial_vs_full} reflect this.  Agents did seem to switch their opinion more often in games with three humans (38\% of the time in full population \textbf{ah} conversations compared to 17\% of the time in partial population \textbf{ah} conversations.)

This change in mean between the full and partial games is larger than the other groups, but it is not a significant difference. It may be possible that agents change switching behavior as a function of the game size, but we do not have enough data to determine such trend with confidence. This result shows that the opinion-switching behavior is largely the same between these full and partial games.

\begin{figure}[t]
    \centering
    \includegraphics[width=0.8\textwidth]{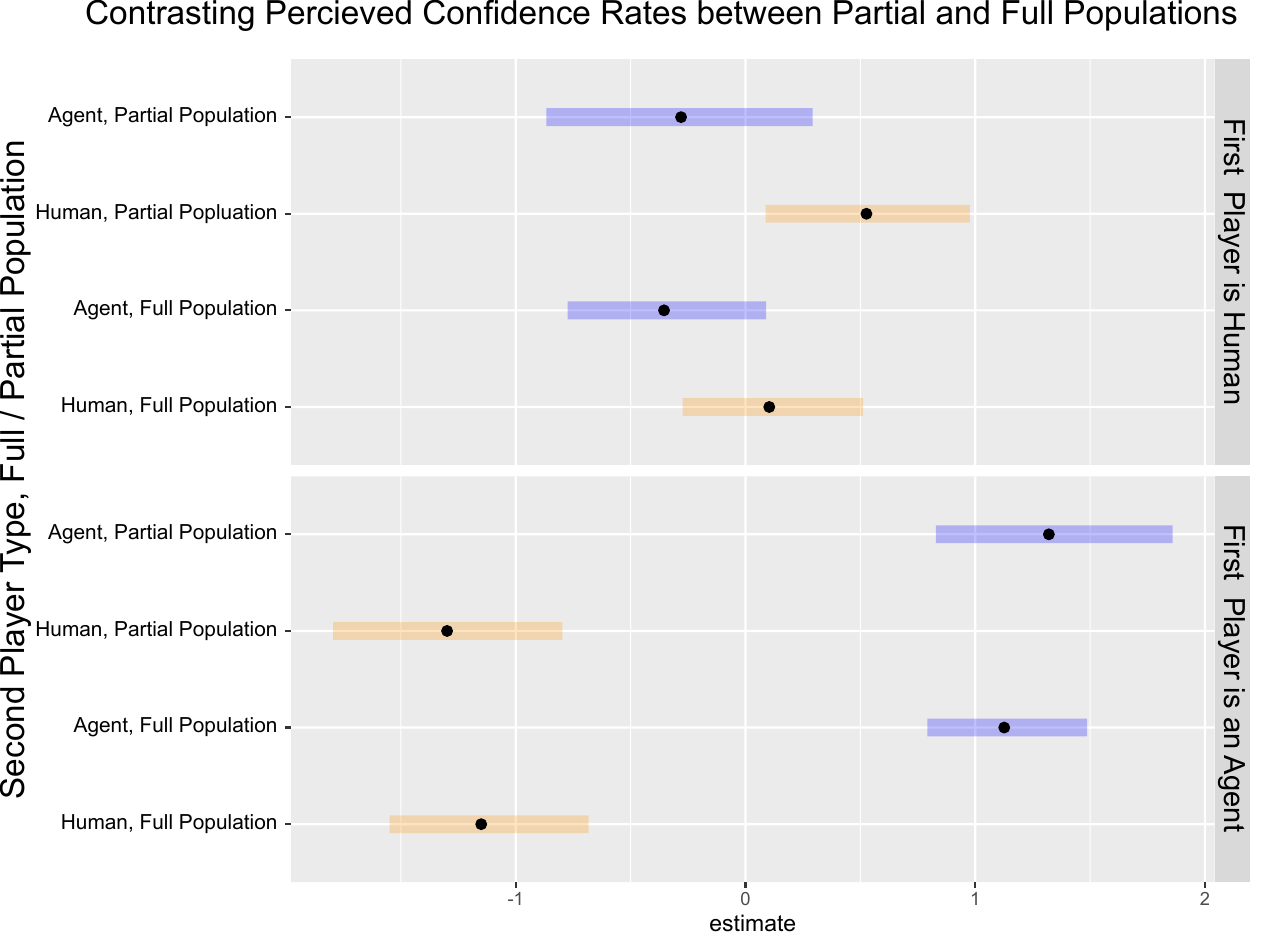}
    \caption{The perceived confidence trends between full and partial games. Each dot and line in the figure indicates the estimate and confidence interval for each combination of Primary participant, secondary participant, and full/partial population game. Each subplot colors the lines based on the second participant. The same two colored lines in each subplot are for full-population games and partial-population games missing a participant. We can see that for each estimate, the means and confidences for the full and partial population games are very similar, indicating no change in perceived confidence behavior when we have partial population games.}
    \label{fig:perceived_conf_partial_vs_full}
\end{figure}

We repeated this analysis for perceived confidence, another important human/agent behavior. We found that humans and agents assigned confidence to their conversational partners in similar ways across both population types, as seen in figure \ref{fig:perceived_conf_partial_vs_full}. Here we see the minimal shifting of the predicted estimates across partial and full population games, indicating the negligible effect of a missing human on game behaviors.

We then compared productivity levels, such as the average length of messages in words, the average number per conversation, and the time between messages. There were more conversations in games with full populations, which is logical given that there were more players to produce conversations. However, conversational markers of productivity remained similar. Based on how similar opinion change, perceived confidence, and productivity measures remained between partial and full population games, we conclude that including partial population games in our overall analyses amplified the signals in our Bayesian analyses, giving us more accurate outcomes without measurable skew.

\subsection{Persona Assignment Procedure}

Our agents had personas that were a combination of four traits, as discussed in detail in the main paper (see ``Agent Design''). Each agent was assigned a persona from a pool of 144 possible personas (with each personas being one unique combination of the four traits). In this section, we describe the procedure followed for agent persona assignments.

First, we create a tabular file with each row containing one persona. Each persona consists of a unique numeric ID between 1 and 144, the values for the four traits associated with that persona, and a record of whether the persona had been assigned to an agent yet. Initially, every persona was marked as 'unused'.

To ensure human-alikeness of our agent personas, we ruled out some unrealistic trait combinations. Specifically, each persona with the 'very confident' initial confidence level and 'suggestible' level of stubbornness was ruled out as unrealistic, and each persona with the 'not very confident' initial confidence level and 'stubborn' level of stubbornness was ruled out as well for the same reason. In this way, 24 personas were ruled out before we played any games.

Then, we used different methods of persona assignment for \textit{AH} and \textit{AA} games. To assign personas for the 51 agents in the 17 \textit{AH} games, we used a random assignment algorithm that generated a random value for each of an agents' four traits, with care taken to avoid the unrealistic personas defined above. It was biased to assign the 'regular' level of stubbornness to agents, as this represents the middle-ground that we hypothesized most humans would also fall under. Due to the nature of random assignment, there was a small probability that some agents between different games would end up with the same persona. As such, we ended up with 30 unique personas between the 51 agents in \textit{AH} games. Those 30 personas were then marked as used in our file.

Next, for the 10 \textit{AA} games, we followed a simple scheme. Each agent was assigned an unused persona from our file, and that persona was marked as used immediately. Thus, each of the 60 agents in the 10 \textit{AA} games had a unique persona. Thus, we have $N_a = 90$ total agents between both game types.

\subsection{Agent Individuation analysis}
Computer agents in sociological experiments tend to be individuated using one or more trait variations. For instance, in [19], the PERSONAGE language generator is used to create agents for human interaction that are individuated within the context of their experiment despite their simplicity when compared to LLMs. GPT-powered LLM agents were used to simulate a small town in [18], wherein the agents were individuated based on their initial traits and relationships with other agents, thus becoming believable human proxies with memory within a specific context. We can also say that choices made by agents individuate them in human studies, as seen in [20], where human infants differentiated between agents based on the choices they saw the agents making.

As stated in the main paper, our study's original count of agents in our study is $N_a = 90$, with 60 agents with 60 unique personas in 10 \textit{AA} games and 51 agents with 30 unique personas in 17 \textit{AH} games. In the following analysis, we compare human and agent individuality. We assign humans a 'personality' by giving them one of three grammar sophistication levels based on their user ID and one of three stubbornness levels based on their initial confidence score. This was the same way we assigned the grammar sophistication level and stubbornness level to our agents.

\begin{table}[]
\begin{tabular}{|l|l|l|l|l|l|l|l|}
\hline
\textbf{Type}  & \textbf{N} & \textbf{1\textsuperscript{st} opinion} & \textbf{Grammar} & \textbf{Stubbornness} & \textbf{1\textsuperscript{st} conf.} & \textbf{2\textsuperscript{nd} opinion} & \textbf{2\textsuperscript{nd} conf.} \\ \hline
Human & 2         & Pescatarian & Simplified   & Regular      & 3                  & Pescatarian & 3, 4                            \\ \hline
Human & 2         & Vegetarian  & Formal       & Regular      & 3                  & Vegetarian  & 3, 2                            \\ \hline
Human & 2         & Omnivorous  & Formal       & Regular      & 3                  & Omnivorous  & 4, 3                            \\ \hline
Human & 2         & Omnivorous  & Simplified   & Stubborn     & 4                  & Omnivorous  & 4, 4                            \\ \hline
Human & 2         & Omnivorous  & Formal       & Regular      & 3                  & Omnivorous  & 3, 3                            \\ \hline
Human & 2         & Omnivorous  & Regular      & Regular      & 3                  & Omnivorous  & 3, 3                            \\ \hline
Human & 2         & Pescatarian & Regular      & Regular      & 2                  & Pescatarian & 2, 0                            \\ \hline
Human & 2         & Vegetarian  & Regular      & Regular      & 3                  & Vegetarian  & 4, 1                            \\ \hline
Agent & 3         & Vegan       & Simplified   & Regular      & 3                  & Vegan       & 4, 2                            \\ \hline
Agent & 2         & Pescatarian & Simplified   & Regular      & 3                  & Pescatarian & 4, 4                            \\ \hline
Agent & 2         & Vegan       & Regular      & Regular      & 3                  & Vegan       & 4, 3                            \\ \hline
Agent & 2         & Vegan       & Regular      & Stubborn     & 4                  & Vegan       & 4, 1                            \\ \hline
Agent & 4         & Vegan       & Regular      & Stubborn     & 4                  & Vegan       & 4, 2                            \\ \hline
Agent & 2         & Vegan       & Simplified   & Stubborn     & 4                  & Vegan       & 4, 2                            \\ \hline
\end{tabular}
\caption{Each pattern of similarity seen among humans and agents. There are 16 humans and 15 agents that share a pattern. The 'N' column shows the number of players that shared that pattern. '1st conf.' shows the initial personal confidence, and '2nd conf.' shows the personal and perceived confidence the player selected after their first conversation.}
\label{tab:similarity_analysis}
\end{table}

We match the following pattern: Initial opinion, stubbornness level, grammar sophistication level, initial confidence, and first conversation results (opinion, personal, and perceived confidence). This pattern shows how unique a participant's personality and decisions were in the first round of the game. Table \ref{tab:similarity_analysis} shows the matching results for agents and humans. We observe that 15 agents and 16 humans had matching patterns, with six different patterns occurring among agents and eight different patterns among humans. Using this metric, agent personality and behavior were at least as unique as human personality and behavior within the first round.

\subsection{Further Individuation Testing}
Our agent individuation analysis aims to test if agent behavior patterns were at least as unique as human behavior patterns within the context of the debate games these agents were designed for. A single behavior pattern, in this case, is the combination of the participant's stubbornness level (suggestible, regular, or stubborn), initial opinion, and pattern of occurrence of a given attribute. An example is an agent with an initial opinion of vegan, stubbornness, and opinion change attribute pattern of vegan-vegetarian-vegan.

\textbf{Agent opinions:} A complete pattern match analysis shows that five pairs of agents across all games had the same stubbornness level, initial opinion, and opinion change pattern. One example of these patterns is Vegan-Vegan-Vegan (two agents had the same stubbornness level, initially a vegan opinion, and had two conversations in the game and did not switch opinions either time).

A partial pattern match analysis was also done, and 34 partial matching pairs were found. In this analysis, if two agents had the same partial pattern, they were counted as a matching pair. (Agent 1 with opinions vegan-vegan and Agent 2 with opinions vegan-vegan-pescatarian would be counted as a match because Agent 1 had two conversations and made the same decisions as Agent 2 did after its first two conversations. This pattern match is independent of several discussions, unlike before).

\textbf{Agent personal and perceived confidence:} We repeat the same analysis as above but with personal confidence. We find that only one pair of agents had a full pattern match, while 21 pairs had a partial pattern match. Finally, for perceived confidence, we found no agent pairs with either full or partial pattern matches.

\textbf{Human opinions:} Humans have ascribed a 'personality' in the same way our agents were so that we could use the exact pattern matching. In other words, humans were given one of three grammar sophistication levels based on their user ID and one of three stubbornness levels based on their initial confidence score. We feel that not assigning a personality to humans in this way (only for this analysis) would be unfair to humans. We would otherwise be comparing their attribute patterns directly without matching a personality first, which would be a more lenient pattern to match. Thus, we will proceed with this approach for comparative analysis of agent and human patterns.

A complete pattern match analysis shows that nine pairs of humans had the same stubbornness level and opinion change behavior. A partial pattern match found 135 matching pairs. This means there are 135 matches, with some humans in multiple pairs.

\textbf{Human personal and perceived confidence:} We repeat the same analysis as above but with personal confidence. We found 11 human pairs with a full pattern match, while 129 pairs had a partial one. Finally, for perceived confidence, we found no matching pairs for either full or partial patterns.

The purpose of this analysis, as stated earlier, was to test if agents were at least as unique in terms of behavioral attribute patterns as humans. We found that humans followed observable patterns more often than agents did across the attributes of opinion change and personal and perceived confidence, as shown by the more extraordinary occurrence of matching behavior pairs among humans compared to agents.

\subsection{Agent individuation and statistical modeling}

In this section we explore the effects of different individuations of agents on the results of our opinion switching model from figure 2 / table 2 of the main text. With this in mind, we explore several options for agent individuation. Each agent could be a "seat" at the table/participant in the study, making each agent conversing with participants with a history provided after each conversation a separate agent. Another interpretation is that each combination of initial opinion and stubbornness level may constitute a different agent. This broader categorization would allow for variations only accounted for by the initialization query to the agent. On the other hand, we can also allow for a single individual agent in our study. These options are in addition to the model we defined in the main paper, where each individual is a unique persona composed of Stubbornness, initial opinion, initial confidence, and grammar type.

We examine how these different individuations of the agents affect our models / explain the variance in the data by performing a model comparison step. In table \ref{sm_tab:agent_individiation_opinion_model_compare}, we outline three variations of the Bayesian Multilevel opinion change model, shown in Table 2 of the main manuscript, each variation one possible individuation of our agents. We selected the One Agent model, the Stubbornness Level x Initial Opinion model, and the All Seats model because they represent a decent spread in the number of individual agents in our study.

\begin{table}[h]
    \centering
    
    \begin{adjustbox}{width=.9\columnwidth,center}
        \begin{tabular}{|rrr|rr|rr|}
          \hline
           & \multicolumn{2}{c|}{One Agent} & \multicolumn{2}{c|}{Stubbornness x Opinion} & \multicolumn{2}{c|}{All Seats}\\
          Predictors & Odds Ratios & CI (95\%) & Odds Ratios & CI (95\%) & Odds Ratios & CI (95\%) \\
          \hline
          \textit{hh} (intercept) & 0.13 & 0.07 - 0.20 & 0.12 & 0.07 - 0.19 & 0.12 & 0.07 - 0.19 \\
          \textit{ha} & 0.16 & 0.05 - 0.46 & 0.16 & 0.04 - 0.46 & 0.16 & 0.04 - 0.45  \\
          \textit{ah} & 3.46 & 1.06 - 15.11 & 3.52 & 1.54 - 8.18 & 2.94 & 1.52 - 5.81 \\
          \textit{aa} & 3.83 & 1.22 - 14.38 & 3.75 & 1.15 - 15.96 & 4.15 & 1.29 - 16.90 \\
          \hline
           \multicolumn{7}{|c|}{Random Effects} \\
           \hline
            Observations & \multicolumn{2}{c|}{1284} & \multicolumn{2}{c|}{1284} & \multicolumn{2}{c|}{1284} \\
            $N_{game}$ & \multicolumn{2}{c|}{37} & \multicolumn{2}{c|}{37} & \multicolumn{2}{c|}{37} \\
            $N_{participant}$ & \multicolumn{2}{c|}{96} & \multicolumn{2}{c|}{107} & \multicolumn{2}{c|}{206} \\
            $\sigma^2$ & \multicolumn{2}{c|}{3.29} & \multicolumn{2}{c|}{3.29} & \multicolumn{2}{c|}{3.29} \\
            $\tau_{00\ game}$ & \multicolumn{2}{c|}{0.16} & \multicolumn{2}{c|}{0.09} & \multicolumn{2}{c|}{0.06} \\
            $\tau_{00\ participant}$ & \multicolumn{2}{c|}{0.35} & \multicolumn{2}{c|}{0.87} & \multicolumn{2}{c|}{0.78}\\
            ICC & \multicolumn{2}{c|}{0.14} & \multicolumn{2}{c|}{0.23} & \multicolumn{2}{c|}{0.20}\\
            Marginal $R^2$ & \multicolumn{2}{c|}{0.042} & \multicolumn{2}{c|}{0.035} & \multicolumn{2}{c|}{0.029} \\
            Conditional $R^2$ & \multicolumn{2}{c|}{0.064} & \multicolumn{2}{c|}{0.149} & \multicolumn{2}{c|}{0.142}\\

           \hline
        \end{tabular}
    \end{adjustbox}
    \caption{\textbf{Bayesian multilevel opinion change models, by agent individuation.} These models outline how the Bayesian multilevel model of opinion change varies as we consider what may constitute an individual agent in our model. We show three such individuation models here, one where there is one agent individual and all agent conversations are its own. The next model is the Stubbornness x Opinion model, which allows for 12 agents, one for each combination of agent stubbornness level and initial opinion. The final model is used in the main text, where each "seat" in a game is its individual. Each model performs largely consistently, with the only exception being that the variance on agent opinion change in \textit{ah} conversations increases as you reduce the number of individual agents. Notably, the conditional $R^2$ for the All Seats and Stubbornness x Opinion model is twice that of the One Agent model. This indicates that the One Seat model explains by far the least variation in the data than when we broadly group them.}
    \label{sm_tab:agent_individiation_opinion_model_compare}

    \end{table}

Table \ref{sm_tab:agent_individiation_opinion_model_compare} shows that these models are similar behaviorally, but the coarser individuations have a higher variance for agent opinion change. The Conditional $R^2$ of the different models captures how much variation in our data is captured by our mixed and random effects. We can see that the Stubbornness x Opinion model explains the most variation, however all but the One Agent model are very close in this regard.

\section{Conversation length and agent budget analysis}

As described in ``Agent design'' in the Methods section, agents in \textit{aa} and \textit{ah} conversations are given a message budget to limit the number of messages they can send and receive. This budget was necessary as the agent's in-built LLM system could not accurately detect if a player sent a message intended to end the conversation. Specifically, after each newly sent message, an auxiliary ChatGPT instance would be prompted with, "You are an AI designed to analyze a message to see if it is a farewell message, parting message, or a message intending to end a conversation. IMPORTANT - You must only respond with a single 'yes' or 'no'. Read the following message: [message content]. Is the message a farewell, parting, or end-of-conversation message? Answer with a single 'yes' or 'no' only." We improve this approach by adding a budget to end conversations in a reasonable time.

\begin{figure}
     \centering
     \begin{subfigure}[b]{0.45\textwidth}
         \centering
         \includegraphics[width=\textwidth]{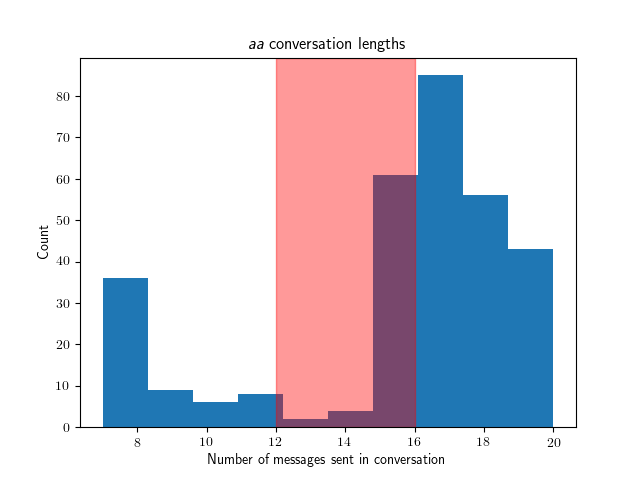}
         \caption{}
         \label{fig:conversation_budget_a}
     \end{subfigure}
     \hfill
     \begin{subfigure}[b]{0.45\textwidth}
         \centering
         \includegraphics[width=\textwidth]{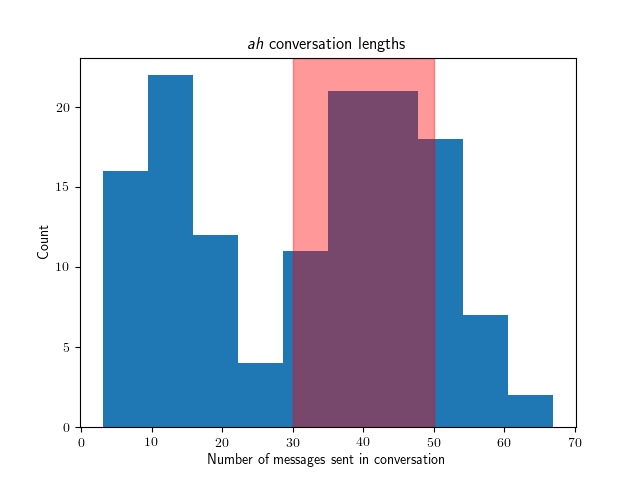}
         \caption{}
         \label{fig:conversation_budget_b}
     \end{subfigure}
     \hfill
     \begin{subfigure}[b]{0.45\textwidth}
         \centering
         \includegraphics[width=\textwidth]{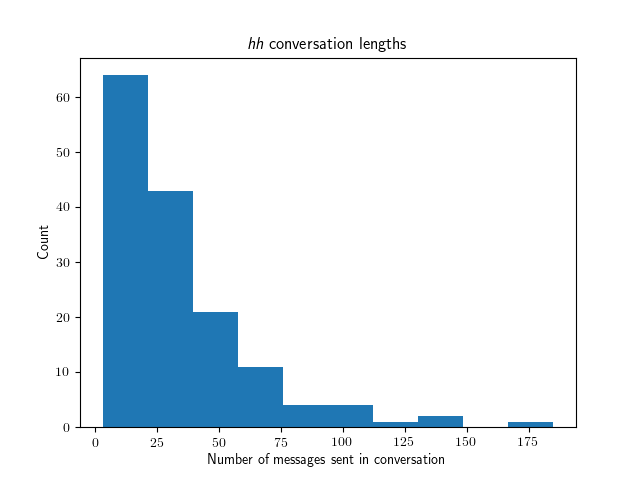}
         \caption{}
         \label{fig:conversation_budget_c}
     \end{subfigure}
        \caption{\textbf{Histograms of conversation lengths by conversation type.} The three plots show the distribution of conversation length in messages for \textit{aa} (\textbf{\subref{fig:conversation_budget_a}}), \textit{ah} (\textbf{\subref{fig:conversation_budget_b}}), and \textit{hh} (\textbf{\subref{fig:conversation_budget_c}}) conversation types respectively. For the \textit{hh} and \textit{ah} plots, the red box indicates the lower and upper bounds of the message budgets allotted to each conversation type. For \textit{aa} the bounds are $(12,16)$ and \textit{ah} are $(30,50)$.}
        \label{fig:conversation_budget}
\end{figure}

Figure~\ref{fig:conversation_budget} shows histograms of conversation lengths by conversation type (including \textit{hh} for comparison). Red boxes are overlaid on the \textit{aa} and \textit{ah} distributions, highlighting the range of the agent message budget defined for each conversation type. Interestingly, these plots show that conversations do not strictly adhere to the limits set by the message budget, even exceeding them occasionally. A case of this can be seen in \textit{ah} conversations (Figure~\ref{fig:conversation_budget}\subref{fig:conversation_budget_b}), with conversations often ending well before the limit, or even ten messages over. Like the \textit{hh} distribution, the \textit{ah} distribution shares a modality near $10$ messages; however, the \textit{ah} distribution also displays a second modality within the budget's bounds. This suggests that human behavior tends toward forming an unimodal distribution with a positive tail (Figure~\ref{fig:conversation_budget}\subref{fig:conversation_budget_c}), but the presence of agents can shift conversation lengths closer to the budget range, producing two modes. The \textit{aa} distribution (Figure~\ref{fig:conversation_budget}\subref{fig:conversation_budget_a}) has a bimodal distribution too, though the second modality dominates. This second modality also forms near the budget range, though not as within bounds as the \textit{ah} distribution. 

As we had found that agents would not terminate the conversation without the budget, we predict that the second modes of the \textit{aa} and \textit{ah} distributions would be positioned farther along the x-axis (as conversations would last much longer), but the modal motifs visible would persist. In the post-game analysis of the main manuscript, we claimed that agents' more static behavior influenced humans in making the conversation amount and length more predictable (relevant to question 1). These results reinforce that observation.

For question 2, we should consider the agents' behavior before budget implementation. Again, as we found that the agents would not correctly terminate conversations without limitations, it is clear that agents are ineffective at producing naturalistic conversation lengths without externally imposed limitations or further refinement. 

\section{Bayesian model definitions}
To determine if there are differences in how humans and agents behave when conversing with each other, we fit Bayesian multi-level models to conversation data between participants. The data used by these models is all the conversations generated over the study, with each conversation providing two data points, one from each participant's perspective on the conversation, with the participant whose opinion we are measuring being $u_1$ and the conversation partner being $u_2$. Each model will, at minimum, include two dependent variables ${u_{1}}\_agent$ and ${u_2}\_agent$, plus their interaction. Both variables are actual that the participant in the conversation was an agent. These two variables and their interaction give us all possible conversation types we wish to model, with \textit{hh} conversations being represented by a 0 for both ${u_{1}}\_agent$ and ${u_2}\_agent$, \textit{ah} conversations with the human being measured is represented by ${u_1}\_agent = 0$ and ${u_2}\_agent = 1$, etc.

The reference level for the model is when both participants are humans (\textit{hh} conversations), and its coefficient is the odds of a human changing their own opinion in an \textit{hh} conversation. The remaining conversation coefficients are shown as the ratio of that conversation's odds and the odds of the reference level, for example, Odds-Ratio$_{ha}=$Odds$_{ha} / $Odds$_{hh}$. 

We use the brms [31] package in R to compute our regression models.

\subsection{Opinion change model}

The opinion change model, presented in Figure 2 and Table 2 of the main manuscript, models the rates at which participants change their opinions in conversations. As described in the previous section, we model this opinion change as a function of the two participants in the discussion and whether they are agents or not.

For reference, the formula used to compute this model in arms is as follows:\
\newline
\
$logit(Odds\ opinion\ change) \sim \beta_0 + \beta_1\ {u_1}\_agent + \beta_2\ {u_2}\_agent + \beta_3\ {u_1}\_agent : {u_2}\_agent$

\
\

Table \ref{sm_tab:bayes_opinion_change_raw} shows us the raw results of our regression model. To relate the model formula to Table \ref{sm_tab:bayes_opinion_change_raw}, \textit{hh} = $\beta_0$, \textit{\textbf{a}h} = $\beta_1$, \textit{a\textbf{h}} = $\beta_2$, and \textit{aa} = $\beta_3$. This differs from the one presented in the main text in how the \textit{aa} row is represented. In the regression model we measure not only the effect of $u_1$ or $u_2$ being an agent separately (which is directly analogous to the Human-Agent (\textit{a\textbf{h}}) conversations and the Agent-Human (\textit{\textbf{a}h}) conversations, but we also have an interaction term between the two variables ${u_{1}}\_agent$ and ${u_2}\_agent$ that measures any additional effect on opinion switching that occurs when both participants are Agents. The \textit{aa} coefficient in this table is a term that modifies the resulting estimate if both participants are Agents, meaning the coefficient alone cannot be used to determine the odds of changing opinion for Agents in \textit{aa} conversations. To compute the Odds or Odds Ratio of the \textit{aa} conversations, we must multiply $\beta_1 \times \beta_2 \times \beta_3$. This is shown in Table 2 of the main text; each row shows the predicted odds of humans and agents changing opinions in each conversation type.

\begin{table}[hp!]
\centering
\begin{adjustbox}{width=.9\columnwidth,center}
\begin{tabular}{|rrr|}
  \hline
  Predictors (conversation type) & Odds (ratios) & CI (95\%)  \\
  \hline
  \textit{hh} (intercept) & 0.12 & 0.07 - 0.19 \\
  \textit{ha} & 0.16 & 0.04 - 0.47 \\
  \textit{ah} & 2.99 & 1.52 - 5.99 \\
  \textit{aa} & 4.09 & 1.27 - 18.06 \\
   \hline
\end{tabular}
\begin{tabular}{|l|l|}
     \hline
    	Observations & 1284 \\
        $N_{game}$ & 37 \\
        $N_{participant}$ & 185 \\
        $\sigma^2$ & 3.29 \\
        $\tau_{00\ game}$ & 0.10 \\
        $\tau_{00\ participant}$ & 0.76 \\
        Marginal $R^2$ & 0.030 \\
        Conditional $R^2$ & 0.140 \\
	\hline
	\end{tabular}
 \end{adjustbox}
 \caption{\textbf{Bayesian multilevel opinion changing model.}
This table defines the parameters and coefficients of this model, presented as odds and odds ratios. The coefficient for \textit{hh} is the model intercept, and its coefficient is the odds that a human changes opinion in \textit{hh} conversations. The remaining coefficients show the odds ratio of opinion change for those conversation types relative to the \textit{hh} odds. Note that the \textit{ah} and \textit{ha} conversation types encompass human-agent interactions where the first letter indicates the participant type whose opinion change is being measured.}

 \label{sm_tab:bayes_opinion_change_raw}
\end{table}

\subsection{Perceived confidence model}
The perceived confidence model, shown in Table \ref{sm_tab:perceived_model}, aims to see the effect of human-agent conversations on how a participant views their conversational partner's confidence. We model perceived confidence as an ordered logit model, with the dependent variables being an interaction term between the participant types of the primary participant ($u_1$) and that of the conversational partner ($u_2$). We model the individual participants and each game as random effects (random intercepts) due to the potential for starting game states (the initial opinions of all participants) affecting perceived confidence and individual factors. 

\begin{table}[htp!]
\centering
\begin{adjustbox}{width=.9\columnwidth,center}
\begin{tabular}{|rrr|}
  \hline
  Predictors & Odds Ratios & CI (95\%)  \\
  \hline
  Perceived confidence 0|1 threshold & 0.02 & 0.01 - 0.03 \\
  Perceived confidence 1|2 threshold & 0.11 & 0.07 - 0.16 \\
  Perceived confidence 2|3 threshold & 0.61 & 0.45 - 0.86 \\
  Perceived confidence 3|4 threshold & 1.37 & 1.00 - 1.91 \\
  \textit{ah} conversations & 0.18 & 0.11 - 0.30 \\
  \textit{ha} conversations & 0.55 & 0.34 - 0.90 \\
  \textit{aa} conversations & 19.44 & 10.01 - 36.96 \\
   \hline
\end{tabular}
\begin{tabular}{|l|l|}
     \hline
        Observations & 1284 \\
        $N_{game}$ & 37 \\
        $N_{participant}$ & 185 \\
        $\sigma^2$ & 3.29 \\
        $\tau_{00\ game}$ & 0.16 \\
        $\tau_{00\ participant}$ & 0.33 \\
        Marginal $R^2$ & 0.148 \\
        Conditional $R^2$ & 0.236 \\
	\hline
	\end{tabular}
 \end{adjustbox}
 \caption{\textbf{Bayesian multi-level perceived confidence model.}
This table defines the parameters and coefficients of this model, with the coefficients presented as odds ratios. Conversation type \textit{ah} indicates agents perceiving human confidence, and \textit{ha} indicates a human perceiving agent confidence. This analysis only includes data from completed conversations. It excludes the 14 participants who didn't actively participate in their game and the two players who conversed for the entire game time without closing the conversation.}
\label{sm_tab:perceived_model}
\end{table}

\subsection{Personal opinion model}
We next model the change in personal confidence as an ordered logistic regression, shown in Table \ref{sm_tab:change_personal_model}. Like the opinion change and perceived confidence models,  the dependent variables are the interaction between the participant types of conversational partners. This model was tested with random effects (random intercepts) on the game and participants. Still, it was determined through model comparison and ICC that these random effects accounted for almost no variance in response, so they were excluded from the final model.

\begin{table}[htp!]
\centering
\begin{adjustbox}{width=.9\columnwidth,center}
\begin{tabular}{|rrr|}
  \hline
  Predictors & Odds Ratios & CI (95\%)  \\
  \hline
  Personal confidence change -3 | -2 threshold & 0.01 & 0.01 - 0.02 \\
  Personal confidence change -2 | -1 threshold & 0.05 & 0.04 - 0.07 \\
  Personal confidence change -1 | 0 threshold & 0.16 & 0.13 - 0.21 \\
  Personal confidence change 0 | +1 threshold & 2.83 & 2.24 - 3.57 \\
  Personal confidence change +1 | +2 threshold & 10.45 & 7.94 - 13.93 \\
  Personal confidence change +2 | +3 threshold & 48.11 & 31.48 - 76.09 \\
  \textit{ah} conversations & 0.76 & 0.50 - 1.15 \\
  \textit{ha} conversations & 0.98 & 0.67 - 1.42 \\
  \textit{aa} conversations & 1.28 & 0.75 - 2.25 \\
   \hline
\end{tabular}
\begin{tabular}{|l|l|}
     \hline
        Observations & 1284 \\
        $N_{game}$ & 37 \\
        $N_{participant}$ & 185 \\
        Conditional $R^2$ & 0.003 \\
        Marginal $R^2$ & 0.005 \\
	\hline
	\end{tabular}
 \end{adjustbox}
 \caption{\textbf{Bayesian multi-level change in personal confidence model.}
This table defines the parameters and coefficients of this model, with the coefficients presented as odds ratios. Conversation type \textit{ha} indicates how human confidence changes when conversing with agents, and \textit{ah} indicates how agent confidence changes when conversing with humans.}
\label{sm_tab:change_personal_model}
\end{table}

\section{Unique word count analysis}

\begin{figure}
     \centering
     \begin{subfigure}[b]{0.45\textwidth}
         \centering
         \includegraphics[width=\textwidth]{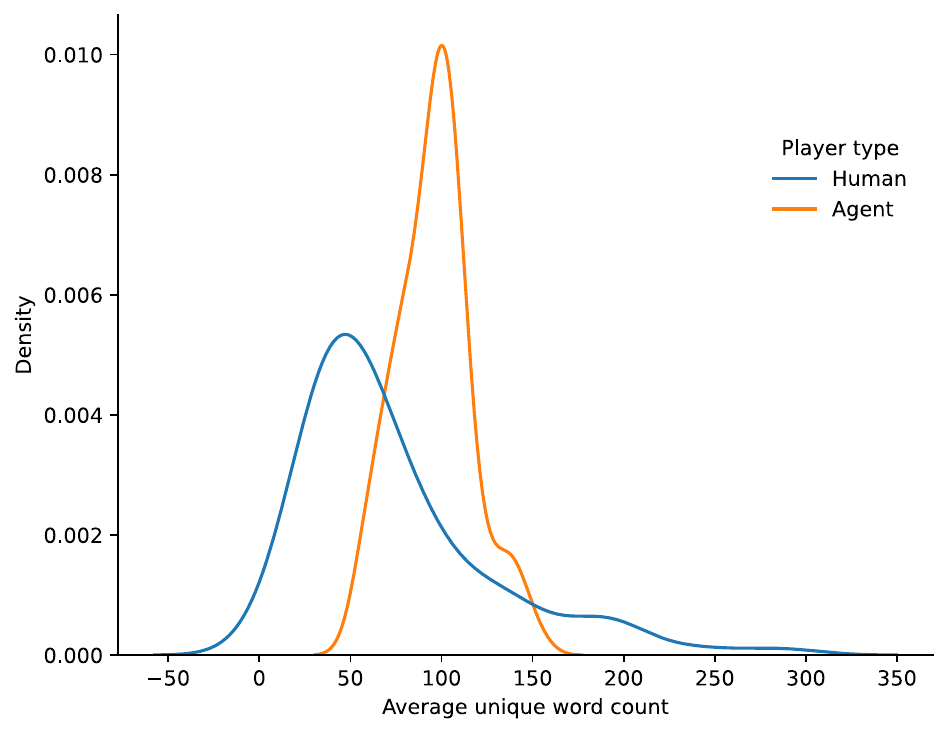}
         \caption{}
         \label{fig:unique_word_count_a}
     \end{subfigure}
     \hfill
     \begin{subfigure}[b]{0.45\textwidth}
         \centering
         \includegraphics[width=\textwidth]{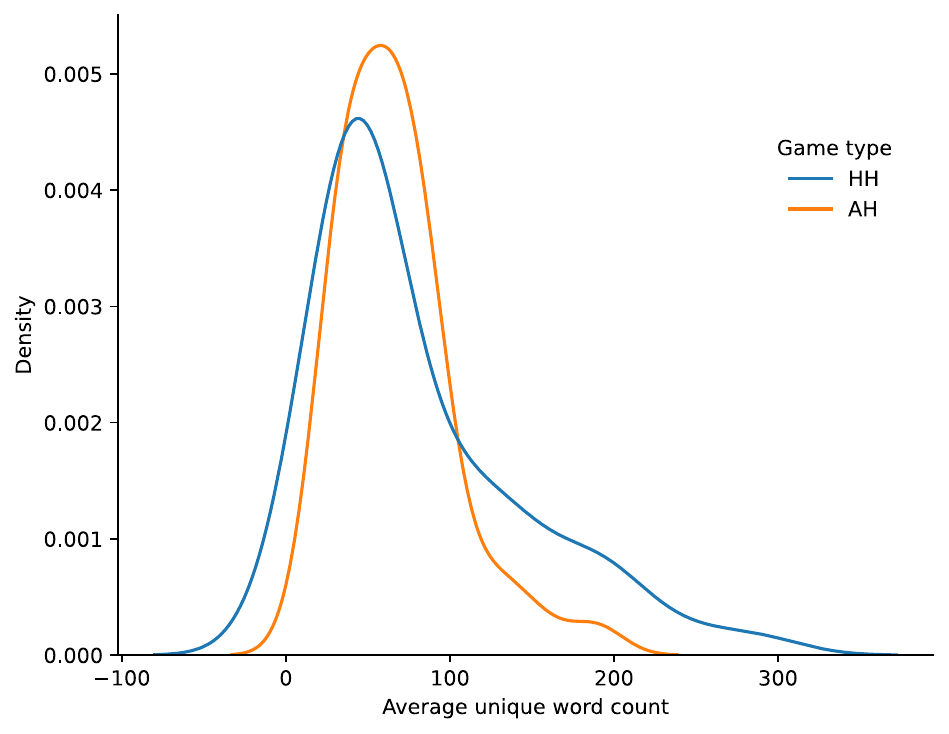}
         \caption{}
         \label{fig:unique_word_count_b}
     \end{subfigure}
        \caption{\textbf{Distributions for unique word count per player.} These distributions show the average number of unique words each player uses in conversations, separated by game type and player type.(\textbf{\subref{fig:unique_word_count_a}}) Kernel Density Estimations (KDEs) of the unique word count for all agents and humans across all game types. (\textbf{\subref{fig:unique_word_count_b}}) KDEs of the unique word count of humans in the \textit{AH} game type versus humans in \textit{HH}.}
        \label{fig:unique_word_count}
\end{figure}

In addition to the on-topic keyword usage of participants, we also analyzed the total number of unique words, excluding commonly used words (i.e., stop words). Here, we repeat all the on-topic keyword analysis done in the main manuscript for unique word counts. Figure~\ref{fig:unique_word_count}\subref{fig:unique_word_count_a} shows the KDEs of the unique word count for all participants, separated by participant type. Figure~\ref{fig:unique_word_count}\subref{fig:unique_word_count_b} shows the KDEs for a unique word count of humans in the \textit{AH} game type compared to those in \textit{HH}. 

Like the on-topic keyword analysis, agents and humans demonstrate notable differences in generating unique words. Agents produce more unique keywords, suggesting their verbosity is intrinsically greater than humans. Unlike the on-topic keyword analysis, however, the comparison of human unique keyword usage in \textit{AH} games versus unique keyword usage in the \textit{HH} game type yielded no significant difference ($p=0.1899$, Welch's t-test).  

\section{Conversational partner selection bias}

During a game, participants can send and accept or reject conversation invites. With the ability to selectively choose what conversations to initiate, the possibility that participants are biased in their selection of conversational partners must be explored. To do this, we consider a $N=6$ graph for a given game, where a node represents a participant and an edge between two nodes represents the two participants accepting a conversation invite, weighted by the number of times a conversation is initiated successfully. From this graph, we can perform community detection to identify potential groupings of participants who converse more consistently with each other than others in a game. Extending this analysis over all \textit{AH} games can allow us to identify if there is a trend in participant type composition of these groups, which would suggest bias in conversational partner selection.

\subsection{Fixed group permutations}

An ideal example of conversational partner selection bias would be in which all humans in an \textit{AH} game prefer conversing with only other humans and the same with agents. This would produce pairs of communities of size $C=3$ each, where each community has a homogeneous participant-type composition. We deploy a brute force approach to see if such community formations are plausible.

This approach consists of generating all permutations of $C=3$ community pairs. A $N=6$ graph yields $10$ community pairs containing all possible community member combinations. We iterate through this set, quantifying the quality of each pair with a connectivity ratio, defined as: \textit{in-group connections $/$ out-group connections}, where a connection is a successfully initiated conversation. Thus, the higher the connectivity ratio for a community, the more intra-community edges there are and the fewer inter-community edges. With this measure, we found the most connected community pair for every \textit{AH} game. The average modularity was $0.0369$ with a standard deviation of $0.0609$, with only 3 of the 17 games yielding a community pair with a homogeneous split in participant-type composition.

The notably low modularity indicates no selection bias in conversational partners. However, to ensure rigor in testing, we explore further a more dynamic community detection approach, which is detailed below. 

\subsection{Louvain community detection}

Using the same graphs as before for the games in the \textit{AH} game type, we used the established Louvain community detection algorithm to identify communities. We analyzed the communities for homogeneous participant-type composition for games with more than one community. Using Louvain community detection, we found $2$ games with $1$ communities, $14$ games with $2$ communities, and $1$ games with $3$ communities. None of these games yielded communities that had homogeneous participant-type composition. Furthermore, the mean modularity of the detected communities was $0.1114$ with a standard deviation of $0.1195$. Ultimately, the results for this and the fixed group permutations suggest no selection bias in conversational partners, as the groups found have very weak modularity. The separation in participant type between groups has no noticeable trend.

\section{Conversation type and game type classifications}

In the main manuscript, our analysis of human and agent behavior showed that a number of our metrics, such as opinion change, conversation length, and on-topic keyword count, have distinguishing trends between game and conversation types. Thus, we ask if these metrics characterize agent and human behavior enough that they can be used to train a simple Machine Learning classifier to differentiate between the \textit{hh} and \textit{ah} conversation types and \textit{HH} and \textit{AH} game types. We consider two classifier models: a Random Forest Classifier (RFC) and a Support Vector Machine (SVM), which we train on $6$ conversational features, allowing us to classify conversations by conversation type and game type. Specifically, for a conversation, we consider on-topic keyword count, unique word count, holding period, response time, total word count, and conversation length (number of messages sent). Notably, the first four features listed are unique to the pair of conversational partners involved, so each conversation has two inputs for these features: one for participant $1$ and one for participant $2$. However, there is no specific order to the participants as it depends on who initiates the conversation; therefore, we combine this dual feature space into one so no classification model arbitrarily values one participant's features over the other. To do this, we implement a pairing function on the features in question, mapping participant $1$'s and participant $2$'s values for each feature into a single value. For the following classification results, we considered both the Szudzik and Cantor pairing functions [29].

\subsection{Conversation type prediction results}

For the conversation type classification problem, we consider all conversations generated within \textit{AH}, producing a dataset of $134$ \textit{ah} conversations and $23$ \textit{hh} conversations. Since this dataset is skewed toward \textit{ah} conversations, we use oversampling to balance the two classes  $15\%$ of the data used for the testing set, with the remainder used for training. We found here that the Szudzik pairing function produced the best results when unifying the values of the on-topic keyword count, unique word count, holding period, and response time features. We also found that the SVM yielded the best results for this classification problem. The best SVM was found to be an RBF kernel where $C=800$ and $\gamma=0.001$. All data was pre-processed using Standard Scaling. With this best-case SVM, the overall accuracy score was $80\%$. Table~\ref{sm_tab:conversation_type_classification_results} shows a more thorough report, including precision, recall, F1, and the SVM's confusion matrix.

\begin{table}[htp!]
\centering
\begin{adjustbox}{width=.9\columnwidth,center}
\begin{tblr}{
  width = \linewidth,
  colspec = {|l|l|l|l|},
  lines,
  lines,
  vline{1} = {1}{},
  vline{3-4} = {-}{},
  hline{1} = {2-4}{},
  hline{2,5} = {-}{},
}
SVM report & \textbf{Precision} & \textbf{Recall} & \textbf{F1-score}\\
\textbf{\textit{ah}} & 1.00 & 0.60 & 0.75\\
\textbf{\textit{hh}} & 0.71 & 1.00 & 0.83\\
\textbf{Weighted averages} & 0.86 & 0.80 & 0.80
\end{tblr}
\begin{tblr}{
  width = \linewidth,
  colspec = {|l|l|l|},
  vline{1-2,4} = {-}{},
  hline{1-2,4} = {-}{},
}
Confusion matrix & \textbf{\textit{ah}} & \textbf{\textit{hh}}\\
\textbf{\textit{ah}} & 12 & 8\\
\hline
\textbf{\textit{hh}} & 0 & 20
\end{tblr}
 \end{adjustbox}
 \caption{\textbf{Report and confusion matrix for conversation type classification.} The table on the left shows the precision, recall, and F1 scores for both labels in this classification problem and for the weighted averages. The table on the right shows the confusion matrix.}
\label{sm_tab:conversation_type_classification_results}
\end{table}

We also performed a feature importance evaluation on the model using permutation importance. The scores produced for each feature used are shown in Table~\ref{sm_tab:feature_importances}. According to the SVM, the most crucial feature for conversation type classification is the total word count, followed by the on-topic keyword count. 

\begin{table}
\centering
\begin{tblr}{
  width = \linewidth,
  colspec = {|l|l|l|},
  cell{2}{2} = {r},
  cell{2}{3} = {r},
  cell{3}{2} = {r},
  cell{3}{3} = {r},
  cell{4}{2} = {r},
  cell{4}{3} = {r},
  cell{5}{2} = {r},
  cell{5}{3} = {r},
  cell{6}{2} = {r},
  cell{6}{3} = {r},
  cell{7}{2} = {r},
  cell{7}{3} = {r},
  lines,
  lines,
}
\textbf{Feature} & \textbf{SVM Score} & \textbf{RF Score}\\
On-topic keyword count & 0.12192982 & 0.313854\\
Unique word count & 0.09078947 & 0.115418\\
Holding period & 0.02280702 & 0.157434\\
Response time & 0.01885965 & 0.102154\\
Total word count & 0.175 & 0.083862\\
Conversation length & 0.01140351 & 0.227278
\end{tblr}
\caption{\textbf{Feature importance scores.} This table shows the feature importance scores computed for the SVM for conversation-type classification and the RF for game-type classification. Permutation importance was used for feature evaluation on the SVM, while for the RF, mean decrease in impurity was used for the RF.}
\label{sm_tab:feature_importances}
\end{table}

\subsection{Game type prediction results}

For our game type classifiers, we consider label prediction on a conversation-by-conversation basis, as there are only $27$ \textit{ah} and \textit{hh} games compared to the $285$ conversations generated within these games, allowing for more training and testing data. Specifically, there are $157$ conversations under \textit{AH} and $128$ under \textit{HH}. Since the two classes are more evenly distributed, we used undersampling to solve the class imbalance. As before, we used $15\%$ of the data for testing. Unlike the previous classification problem, we found that an RFC with ten estimators with two max features and unbounded depth yielded the best results. Furthermore, we found that the Cantor pairing performed better than Szudzik here. Overall, the accuracy of this setup was $71\%$. See Table~\ref{sm_tab:condition_classification_results} for more details on prediction performance.

\begin{table}[htp!]
\centering
\begin{adjustbox}{width=.9\columnwidth,center}
\begin{tblr}{
  width = \linewidth,
  colspec = {|l|l|l|l|},
  lines,
  lines,
  vline{1} = {1}{},
  vline{3-4} = {-}{},
  hline{1} = {2-4}{},
  hline{2,5} = {-}{},
}
RF report & \textbf{Precision} & \textbf{Recall} & \textbf{F1-score}\\
\textbf{\textit{AH}} & 0.68 & 0.72 & 0.70\\
\textbf{\textit{HH}} & 0.74 & 0.70 & 0.72\\
\textbf{Weighted averages} & 0.71 & 0.71 & 0.71
\end{tblr}
\begin{tblr}{
  width = \linewidth,
  colspec = {|l|l|l|},
  vline{1-2,4} = {-}{},
  hline{1-2,4} = {-}{},
}
Confusion matrix & \textbf{\textit{AH}} & \textbf{\textit{HH}}\\
\textbf{\textit{AH}} & 13 & 5\\
\hline
\textbf{\textit{HH}} & 6 & 14
\end{tblr}
 \end{adjustbox}
 \caption{\textbf{Report and confusion matrix for game type classification.} The table on the left shows the precision, recall, and F1 scores for both labels in this classification problem and for the weighted averages. The table on the right shows the confusion matrix.}
\label{sm_tab:condition_classification_results}
\end{table}

We also evaluated the features for the RF used in game type classification, presented in Table~\ref{sm_tab:feature_importances}. Here, feature importance was scored through a mean decrease in impurity. In Table~\ref{sm_tab:feature_importances}, on-topic keyword count is shown to be the most crucial feature of the RFC in classifying game types, followed by conversation length.

\subsection{Additional discussion}

The above results indicate that a set of conversational features identified in the main manuscript enables simpler classifier models to differentiate between \textit{ah} and \textit{hh} conversations and \textit{AH} and \textit{HH} game types. Furthermore, the classifier models' accuracy suggests that the two classes' behavior diverges significantly. Overall, we can consider these observations in the context of the main manuscript's question $2$: can generative LLM simulations produce realistic, human-like behavior in conversation-intensive studies? In particular, we find in this analysis that behavior relating to on-topic keyword count, total word count, and conversation length distinctly characterizes the participant type, with on-topic keyword count as an essential feature in both classification problems. Therefore, we posit that keyword usage is one of the primary attributes in agent behavior that differentiates them from humans.

\section{Initiator analysis and inviter-invitee dynamics}

Another avenue of exploration that must be covered is an analysis of conversation types based on the inviter and invitee. In our study's games, all participants can send invites to and accept them from other participants to join a one-on-one conversation. Since a participant sending out an invite is taking the initiative to push their opinion onto another participant, there might be a difference in behavior between the two involved. Agents can send out invites using a scripted Python routine. When they are not conversing, agents idle for a few minutes, waiting to receive invitations from other participants. If no invitations are accepted, agents invite the first available participant without bias toward humans or agents.

\subsection{Tests for \textit{ah} conversations}

To start our analysis, we ask if humans in \textit{ah} conversations have different behaviors depending on whether they invited the agent or if the agent invited them. In the former case, the human would be the inviter (or initiator), the agent would be the invitee (or non-initiator), and vice-versa, giving us two conversation groupings. To analyze this, we consider four characteristics for each participant: unique word count (UWC), on-topic keyword count (OTKC), holding period (HP), and response time (RT). We also consider two conversation-wide characteristics: word count (WC), the total number of words in the conversation, and conversation length (CL), the total number of messages sent and received. This gives us six characteristics. 

To test whether behavior dynamics were significantly different between these two groups, we propose a null hypothesis that can be applied to all six attributes.

\textbf{Null hypothesis:} For a given attribute, there is no statistically significant difference in that attribute for humans between human-initiated and agent-initiated \textit{ah} conversations.

\begin{table*}[h]
    \centering
    \renewcommand{\arraystretch}{1.25} 
    \begin{tabular}{|g|*{6}{c|}}
        \hline
        \rowcolor{lightgray}
         & \textbf{UWC} & \textbf{OTKC} & \textbf{HP} & \textbf{RT} & \textbf{WC} & \textbf{CL} \\
        \hline
        \textbf{Human-Initiated} & 66.80 & 15.63 & 59.13 & 167.37 & 1,647.22 & 31.18 \\
        \hline
        \textbf{Agent-Initiated} & 45.52 & 8.95 & 53.02 & 169.81 & 1,202.62 & 22.81 \\
        \hline
    \end{tabular}
    \caption{Mean values for the six chosen conversation characteristics for humans when the conversation is human-initiated and when it is agent-initiated.}
    \label{tab:agent-human initiator values} 
\end{table*}

\begin{table*}[h]
    \centering
    \renewcommand{\arraystretch}{1.25} 
    \begin{tabular}{|g|*{6}{c|}}
        \hline
        \rowcolor{lightgray}
         & \textbf{UWC} & \textbf{OTKC} & \textbf{HP} & \textbf{RT} & \textbf{WC} & \textbf{CL} \\
        \hline
        \textbf{p-value} & \textbf{0.01} & \textbf{0.001} & 0.67 & 0.94 & \textbf{0.02} & \textbf{0.007} \\
        \hline
    \end{tabular}
    \caption{Independent two-sample t-test outcomes for each attribute. Bold values indicate a statistically significant difference in those attributes of human behavior between human and agent-initiated conversations. Values not bold indicate that the null hypothesis was accepted.}
    \label{tab:agent-human t-tests} 
\end{table*}

We take the entire populations of these attributes across the two groupings and perform independent two-sample t-tests. As before, we calculate population variance and perform Welch's t-test if variances are unequal. The average values for each attribute for the two groups are in table \ref{tab:agent-human initiator values}. The results of the t-tests can be seen in table \ref{tab:agent-human t-tests}.

The results above tell us that humans tend to be more active in conversations that they initiate. This increase in activity is consistent for five of the six attributes and is statistically significant for four. The most important difference is for on-topic keyword counts followed by conversation length. We may interpret from these outcomes that humans were more focused and motivated when initiating an \textit{ah} conversation. However, holding periods and response times remain comparable between human initiators and non-initiators. This may be because the time spent sending a chain of messages and response speed does not depend on how motivated a participant was. Overall, conversation initiation appears to be an essential factor in conversation activity levels.

The question of why we see this trend is more difficult to answer. One may suggest that agents are poor conversation initiators, negatively impacting human productivity because their openings are not good enough. However, this is unlikely because agents begin conversations with casual greetings similar to humans. This may be because productive human participants are highly motivated and, thus, more likely to be initiators. We test this theory in the following subsection.

\subsection{Inviter-invitee productivity comparison}

If human initiators are more productive because productive participants are more likely to initiate conversations, then initiators in \textit{hh} conversations should show the same trends as those in \textit{ah} conversations. We test this by comparing initiators and non-initiators across all \textit{HH} games as well as \textit{hh} conversations from \textit{AH} games. We test the four attributes each participant can uniquely have (OTWK, UWC, HP, RT) but not word count or conversation length. This is because we test all conversations instead of dividing them into two sets (since dividing \textit{hh} conversations based on initiator type is impossible). Thus, our null hypothesis changes slightly.

\textbf{Null hypothesis:} For a given attribute, there is no statistically significant difference in that attribute between initiators and non-initiators.

\begin{table*}[h]
    \centering
    \renewcommand{\arraystretch}{1.25} 
    \begin{tabular}{|g|*{4}{c|}}
        \hline
        \rowcolor{lightgray}
         & \textbf{UWC} & \textbf{OTKC} & \textbf{HP} & \textbf{RT} \\
        \hline
        \textbf{\textit{hh} from \textit{HH} p-value} & 0.44 & 0.60 & 0.40 & 0.65 \\
        \hline
        \textbf{\textit{hh} from \textit{AH} p-value} & 0.76 & 0.81 & 0.33 & 0.76 \\
        \hline
    \end{tabular}
    \caption{Independent two-sample t-test outcomes for each attribute for \textit{hh} conversations from \textit{HH} games and for \textit{hh} conversations from \textit{AH} games.}
    \label{tab:human-only t-tests} 
\end{table*}

As can be seen from the p-values reported in table \ref{tab:human-only t-tests}, the behavior of humans that initiate conversations does not differ significantly from that of humans on the receiving end. This implies that initiating a conversation only makes humans more productive in agent-human interaction. Further research is required to determine the exact cause of this difference.

\subsection{Opinion changing behaviors of initiators vs non-initiators}

Another open question is how initiating a conversation affects participants' tendency to change their opinions. Table \ref{tab:initiator_opinion_change} shows the counts of participants changing opinions and whether or not those participants initiated the conversation. We can see in the \textit{HH} row that $\sim7\%$ of humans who initiated conversations switched opinions and $\sim18\%$ of humans who didn't initiate switched. This hints that a relationship may exist between initiating a conversation and a human's decision to switch their opinion. We analyze this trend further by modeling the decision to switch views for \textit{HH} games as a function of one dichotomous variable: whether the participant who switched was the conversation initiator. As with our other regression models, we add participants and games as random effects to ensure that game and participant-specific variances don't skew the results.

\begin{table}[ht!]
    \centering
    \begin{tabular}{|g|g|g|g|g|g|}
    \hline
     &  & \multicolumn{4}{|g|}{\textbf{Did the participant change opinion?}} \\
     &  &  \multicolumn{2}{g}{\textbf{Yes}} &  \multicolumn{2}{g|}{\textbf{No}}\\ \cline{3-6}
    \multirow{-3}{*}{\textbf{Population}} & \multirow{-3}{*}{\textbf{Conversation type}} &  Initiated &  Didn't Initiate & Initiated &  Didn't Initiate\\ \hline
    \textit{HH} & \textit{hh} & 9 & 23 & 114 & 104\\ \hline
    \textit{AA} & \textit{aa} & 57 & 68 & 178 & 167\\ \hline
     & \textit{hh} & 5 & 2 & 17 & 21\\ 
     & \textit{ha} & 4 & 0 & 88 & 42\\ 
    \multirow{-2}{*}{\textit{AH}} & \textit{ah} & 13 & 30 & 29 & 63\\ 
     & \textit{aa} & 20 & 18 & 54 & 56\\ \hline
    \end{tabular}
    \caption{\textbf{Summary of Initiator Opinion Changes, all games.} This table contains the opinion change outcome of all conversations across all three populations, broken out by participants who did and did not initiate the conversation. As in the main manuscript, we split conversation type \textit{ah} into \textit{ah} and \textit{ha} to delineate the opinion change of the agent and the human involved in the conversation, respectively. Incomplete conversations do not yield any opinion change, and there is an instance where one participant finished re-evaluating their opinion before the time limit expired, but their conversational partner did not.} 
    \label{tab:initiator_opinion_change}
\end{table}

We can see from the results of the Bayesian Regression model, Table \ref{tab:hh_initiator_regression}, that initiating the conversation in \textit{HH} games makes a participant $1/3$ as likely to switch their opinion as non-initiators. This effect is significant, as evidenced by the $95\%$ CI not overlapping with 1.

\begin{table}[htp!]
\centering
\begin{adjustbox}{width=.9\columnwidth,center}
\begin{tabular}{|rrr|}
  \hline
  Predictors & Odds ratios & CI (95\%)  \\
  \hline
  Did not Initate (intercept) & 0.19 & 0.09 - 0.37 \\
  Initiated & 0.32 & 0.13 - 0.76 \\
   \hline
\end{tabular}
\begin{tabular}{|l|l|}
     \hline
    	Observations & 250 \\
        $N_{game}$ & 10 \\
        $N_{participant}$ & 52 \\
        $\sigma^2$ & 3.29 \\
        $\tau_{00\ game}$ & 0.29 \\
        $\tau_{00\ participant}$ & 0.32 \\
        Marginal $R^2$ & 0.022 \\
        Conditional $R^2$ & 0.081 \\
	\hline
	\end{tabular}
 \end{adjustbox}
 \caption{\textbf{Initiator opinion change model, \textit{hh} conversations.}
This table defines the parameters and coefficients of this model. The coefficient of the intercept is the odds that a non-initiator of a \textit{hh} conversation in an \textit{HH} game switched their opinion. The coefficient for Initiated shows the odds ratio for the initiators of conversations versus the non-initiators.}

 \label{tab:hh_initiator_regression}
\end{table}

We next examine how the dynamics of opinion switching changed in \textit{ah} conversations depending on whether agents versus humans initiated the conversation. Table \ref{tab:initiator_counts} shows these counts of initiators of \textit{ah} conversations by agents versus humans, as well as the opinion-switching counts for those who initiated and didn't initiate conversations. From this table we can see that the rates of opinion switching seem very nearly the same between initiators and non-initiators for both agents and humans, with agents having only a $\sim2\%$ difference in switching frequency between initiators and non-initiators and humans having a $\sim4\%$ difference.

\begin{table*}[h]
    \centering
    \renewcommand{\arraystretch}{1.25}
    \begin{subtable}[t]{0.45\textwidth}
    \begin{tabular}{|g|*{2}{c|}}
        \hline
        \rowcolor{lightgray}
         & \textbf{Human} & \textbf{Agent}   \\
        \hline
        \textbf{Didn't Initiate} & 42 & 93 \\
        \hline
        \textbf{Initiated} & 92 & 42 \\
        \hline
    \end{tabular}%
    \caption{\textbf{Initiators counts by Human / Agent.} }
    \label{tab:initiator_counts_a}
    \end{subtable}
    
    \begin{subtable}[t]{0.45\textwidth}
    \begin{tabular}{|g|*{2}{c|}}
        \hline
        \rowcolor{lightgray}
        \textit{Initiated} & \textbf{Human} & \textbf{Agent}   \\
        \hline
        \textbf{Stayed} & $88_{(.956)}$ & $29_{(.690)}$  \\
        \hline
        \textbf{Switched} & $4_{(.043)}$ & $13_{(.309)}$ \\
        \hline
    \end{tabular}
    \caption{\textbf{Initiator Opinion switch counts.} }
    \label{tab:initiator_counts_b}
    \end{subtable}
    \begin{subtable}[t]{0.45\textwidth}
    \begin{tabular}{|g|*{2}{c|}}
        \hline
        \rowcolor{lightgray}
        \textit{Didn't Initiate} & \textbf{Human} & \textbf{Agent}   \\
        \hline
        \textbf{Stayed} & $42_{(1.0)}$ & $63_{(.677)}$  \\
        \hline
        \textbf{Switched} & $0_{(0.0)}$ & $30_{(.322)}$ \\
        \hline
    \end{tabular}
    \caption{\textbf{Non-Initiator Opinion switch counts in \textit{ah} conversations.} }
    \label{tab:initiator_counts_c}
    \end{subtable}
    \caption{\textbf{Initiators and opinion switching counts.} These tables show the counts of participants who did and did not initiate conversations and those who switched while being initiators and non-initiators.}
    \label{tab:initiator_counts}
\end{table*}

To determine if the human switching differences in \textit{ah} conversations are significant, we use a slightly modified version of the previous model (Table \ref{tab:hh_initiator_regression}) where we include an additional dichotomous variable for if the participant switching opinion is a human or an agent and we include a term for the interaction of initiating the conversation and being a human/agent. This allows us to model how initiating the conversation affects humans' and agents' decisions to switch opinions. We attempted to run this model on \textit{ah} conversation data, but the results indicate insufficient data points on \textit{ah} conversations alone to detect any effect. The model ran into various issues that point towards inadequate data points and extreme over-fitting (high numbers of divergent chains, hyper-inflated log odds, model identifiability issues, etc.). Due to this issue, we cannot draw any conclusions on the effect of initiating conversations and opinion switching in $ah$ conversations.

\section{Participant preparation for the game, essential compensations, and additional awards}

The primary compensation for participants is USD \$10 (in the form of an Amazon gift card) for completing a full game lasting 60 minutes. Furthermore, these games involve six participants, and each participant reads the open-ended prompt and selects one of four different opinions on this prompt. Notably, the actual prompt is not revealed to any participants until they enter a game. The rules also describe the game's goal as finishing with the most points earned by convincing other participants to change their opinion to theirs and by finding the majority opinion by the game's conclusion (worth one and three points, respectively). Participants were instructed that, at the end of the game, the top two participants, in terms of points gained, would receive an additional USD \$10 on top of their base compensation (the value of which is appended to their Amazon gift card). The rules then explain that all participants must take the exit survey to provide their payment details and other post-game information. Finally, participants are informed that after completing the game and exit survey, they are emailed their compensation and permanently removed from the study.

For instructions on how to join a game, participants are directed to a scheduling system integrated into the Discord channel using Discord's bot API. This Discord bot will periodically post a series of game times a day in advance. Participants will anonymously indicate their availability during these game times. Importantly, they can mark availability as many times as they see fit. Once the minimum required number of available participants has been reached (three for an \textit{AH} game and six for a \textit{HH} game), the game time itself is announced to the participant an hour before the game starts by the Discord bot. Suppose the number of available participants at a given time is greater than the minimum required for the type of game being played. In that case, the Discord bot will select a random subset of the available participants to play. One minute before the game, the Discord bot gives the chosen participants their login credentials (a randomly generated password and one of six preset usernames) and a link to the game user interface.

The rules explain that participants are compensated USD \$10 (in the form of an Amazon gift card) for participating in a complete game lasting 60 minutes. Furthermore, these games involve six participants, and each participant must read the open-ended prompt and select one of four different opinions on this prompt. Notably, the actual prompt is not revealed to any participants until they enter a game. The rules also describe the game’s goal as finishing with the most points earned by convincing other participants to change their opinion to theirs and by finding the majority opinion by the game’s conclusion (worth one and three points, respectively). Participants are instructed explicitly that, at the end of the game, the top two participants, in terms of points gained, will receive an additional USD \$10 on top of their base compensation (the value of which is appended to their Amazon gift card). The rules then explain that all participants must take the exit survey to provide their payment details and other post-game information. Finally, participants are informed that after completing the game and exit survey, they are emailed their compensation and permanently removed from the study.

For instructions on how to join a game, participants are directed to a scheduling system integrated into the Discord channel using Discord’s bot API. This Discord bot will periodically post a series of game times a day in advance. Participants will anonymously indicate their availability during these game times. Importantly, they can mark availability as many times as they see fit. Once the minimum required number of available participants has been reached (three for an \textit{AH} game and six for a \textit{HH} game), and the game time itself is an hour away, the Discord bot will inform the participants that the game is occurring later in the day. Suppose the number of available participants at a given time is greater than the minimum required for the type of game being played. In that case, the Discord bot will select a random subset of the available participants to play. One minute before the game, the Discord bot gives the chosen participants their login credentials (a randomly generated password and one of six preset usernames) and a link to the game user interface.

\newpage

\section{Majority Consensus Analysis}

We analyze the majority consensus groups in the entire population \textbf{AH} games to see where each type of player stood. Convincing other players to change their opinions and being in the opinion group with a majority consensus were the primary goals that each player was incentivized to achieve.

\begin{table}[h]
\begin{tabular}{|l|l|l|l|l|}
\hline
\textbf{S. No.} & \textbf{Consensus size} & \textbf{Population types} & \textbf{Majority opinion} & \textbf{Eliminated opinion(s)} \\ \hline
1               & 5                       & 3A, 2H                    & VE                        & V, P                           \\ \hline
\rowcolor[HTML]{EFEFEF} 
2               & 3                       & 1A, 2H                    & V                         & None                           \\ \hline
3               & 3; 3                    & 3A;3H                    & V, P                      & VE, O                          \\ \hline
\rowcolor[HTML]{EFEFEF} 
4               & 3                       & 1A, 2H                    & V                         & P                              \\ \hline
5               & 3                       & 2A, 1H                    & V                         & O                              \\ \hline
\rowcolor[HTML]{EFEFEF} 
6               & 4                       & 2A, 2H                    & P                         & VE, O                          \\ \hline
7               & 3                       & 1A, 2H                    & O                         & VE                             \\ \hline
\rowcolor[HTML]{EFEFEF} 
8               & 5                       & 2A, 3H                    & P                         & VE, V                          \\ \hline
9               & 3                       & 2A, 1H                    & V                         & VE                             \\ \hline
\rowcolor[HTML]{EFEFEF} 
10              & 5                       & 2A, 3H                    & P                         & VE, V                          \\ \hline
\end{tabular}
\caption{The majority consensus groups with a full population at the end of each AH game. 'Population types' shows the breakdown of participant types (A-Agent, H-Human) in the consensus groups. 'Majority opinion' shows the diet agreed upon by the majority (VE-Vegan, V-Vegetarian, P-Pescatarian, O-Omnivorous). 'Eliminated opinion(s)' shows the diets that no one held at the end of the game. Game No. 3 had two equal majority consensus groups.}
\label{tab:consensus_groups}
\end{table}

Table \ref{tab:consensus_groups} shows the final consensus groups, including the size of the groups, the diet represented, the numbers of each participant type holding consensus, and the diets eliminated from contention. Six games had a consensus group size of 3, one size of four, and three sizes of 5. Only one game had two equal majority consensus groups of size 3. We observe that agents and humans are represented similarly in the table, with 19 of 30 agents and 21 of 30 humans in their respective majority consensus groups. This analysis shows that both participant types could reach group consensus successfully.

Additionally, we found that in \textbf{HH} games, The most popular opinion was Omnivorous, followed by Pescatarian. In \textbf{AA} games, Vegan was the most popular opinion, followed by Vegetarian and Pescatarian. In \textbf{AH} games, Pescatarian and Vegetarian were the most popular, followed by the other two.

\newpage

\section{Additional figures and tables}

\begin{figure}[htp!]
    \centering
    \includegraphics[width=0.32\linewidth]{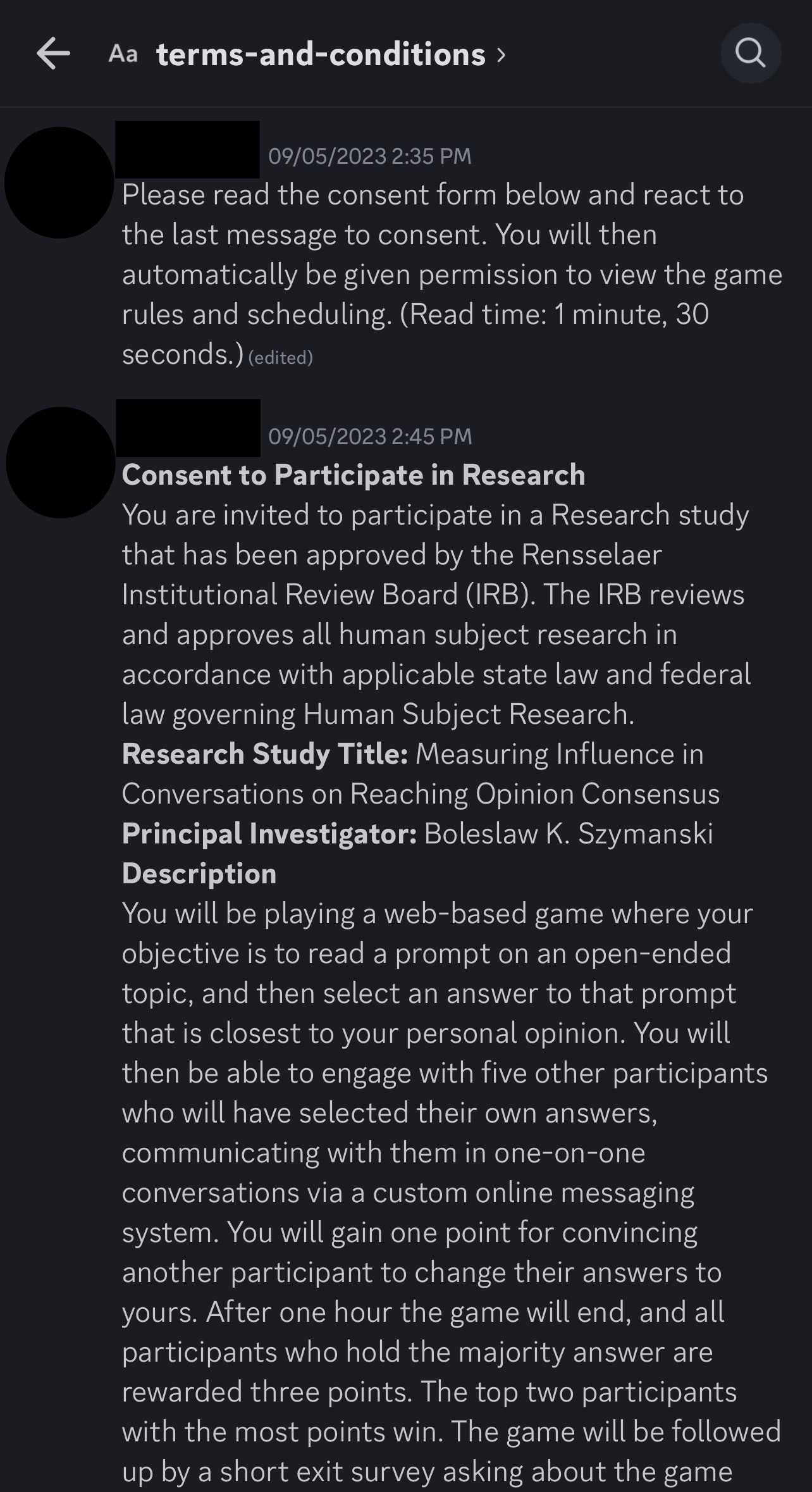}
    \includegraphics[width=0.32\linewidth]{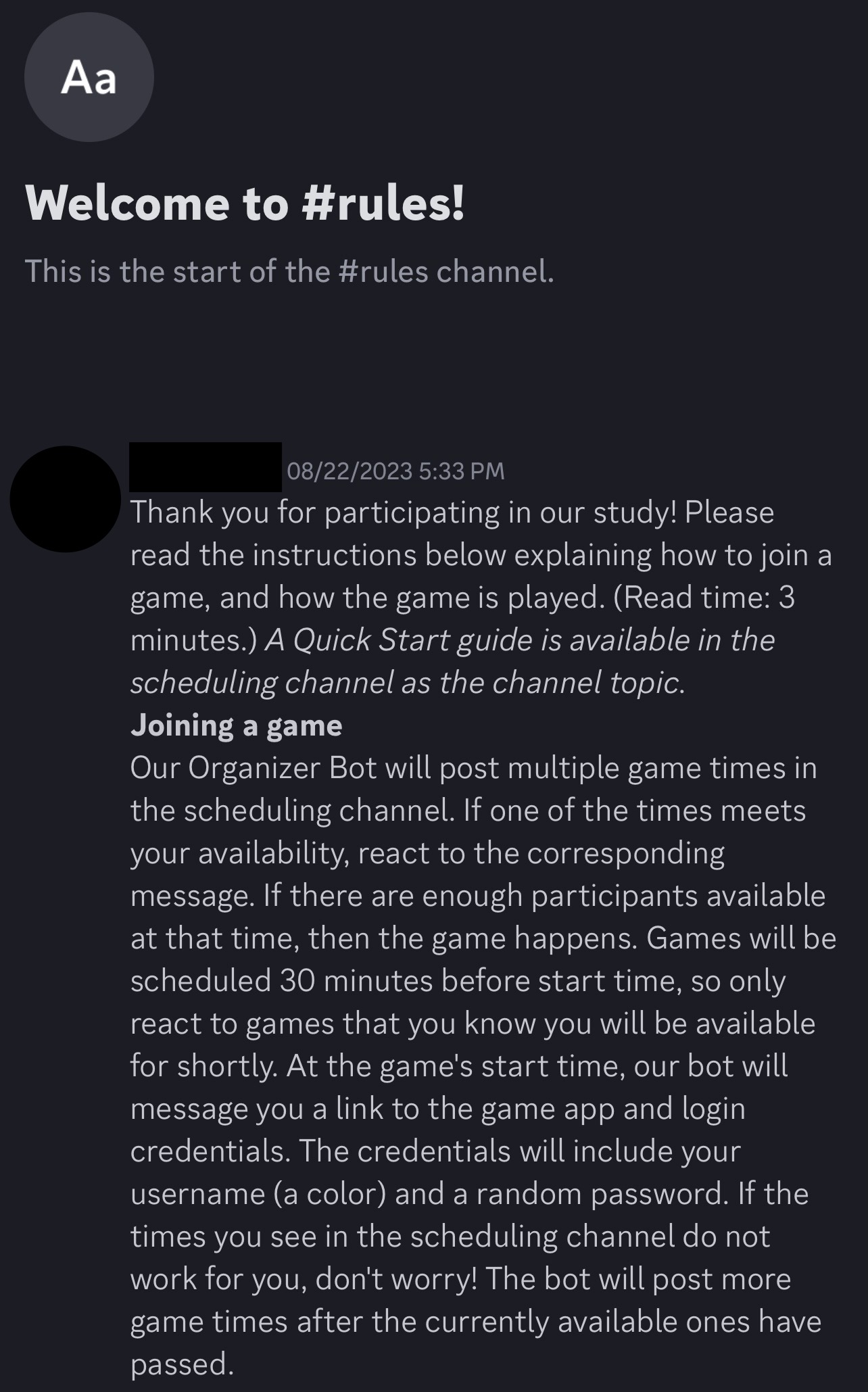}
    \includegraphics[width=0.32\linewidth]{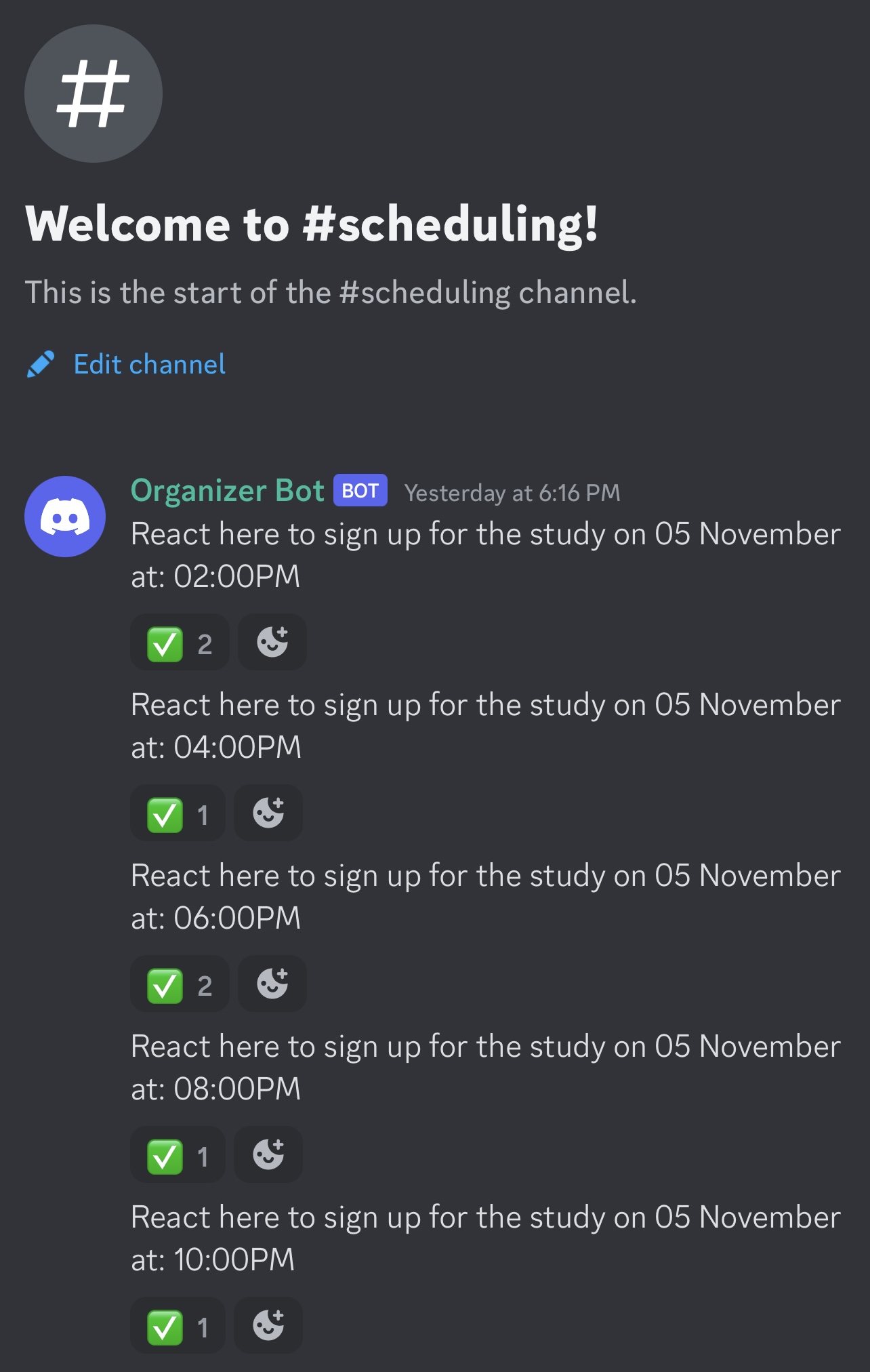}
    \caption{\textbf{Screenshots of the study's Discord channel.} The left screenshot shows the terms and conditions. This is the first page of the Discord that participants must see. At the end of this page, participants must indicate they consent to the research and are at least 18 years of age. After this, they are directed to the rules and scheduling pages (middle and right screenshots). Our Discord agent generated scheduling messages, but a moderator and a research team member produced all other messages. The moderator's Discord account has been censored for their privacy.}
    \label{fig:discord_screenshots}
\end{figure}

\begin{figure}[htp!]
    \centering
    \includegraphics[width=0.49\linewidth]{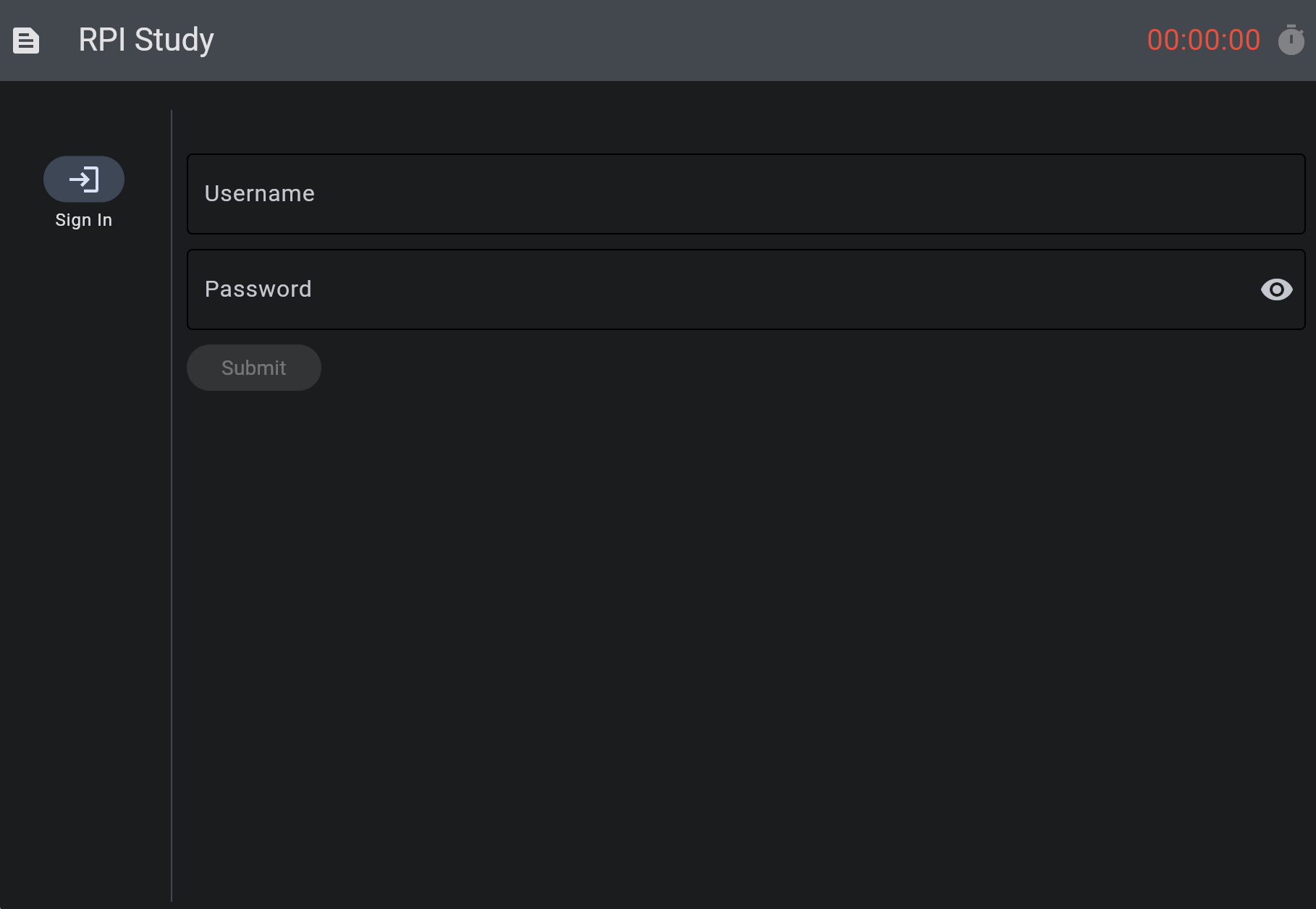}
    \includegraphics[width=0.49\linewidth]{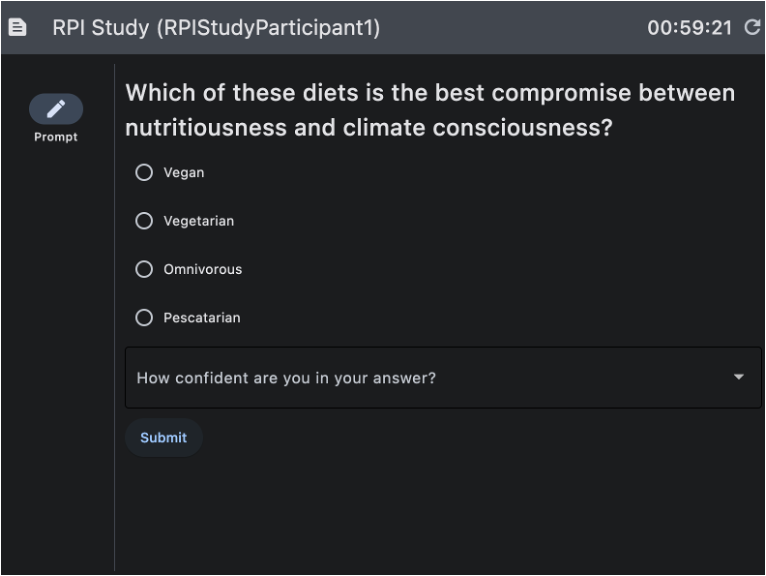}
    \includegraphics[width=0.49\linewidth]{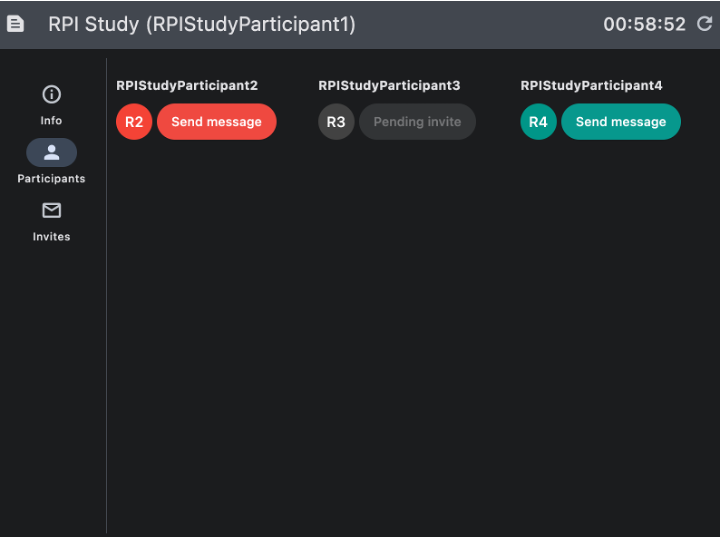}
    \includegraphics[width=0.49\linewidth]{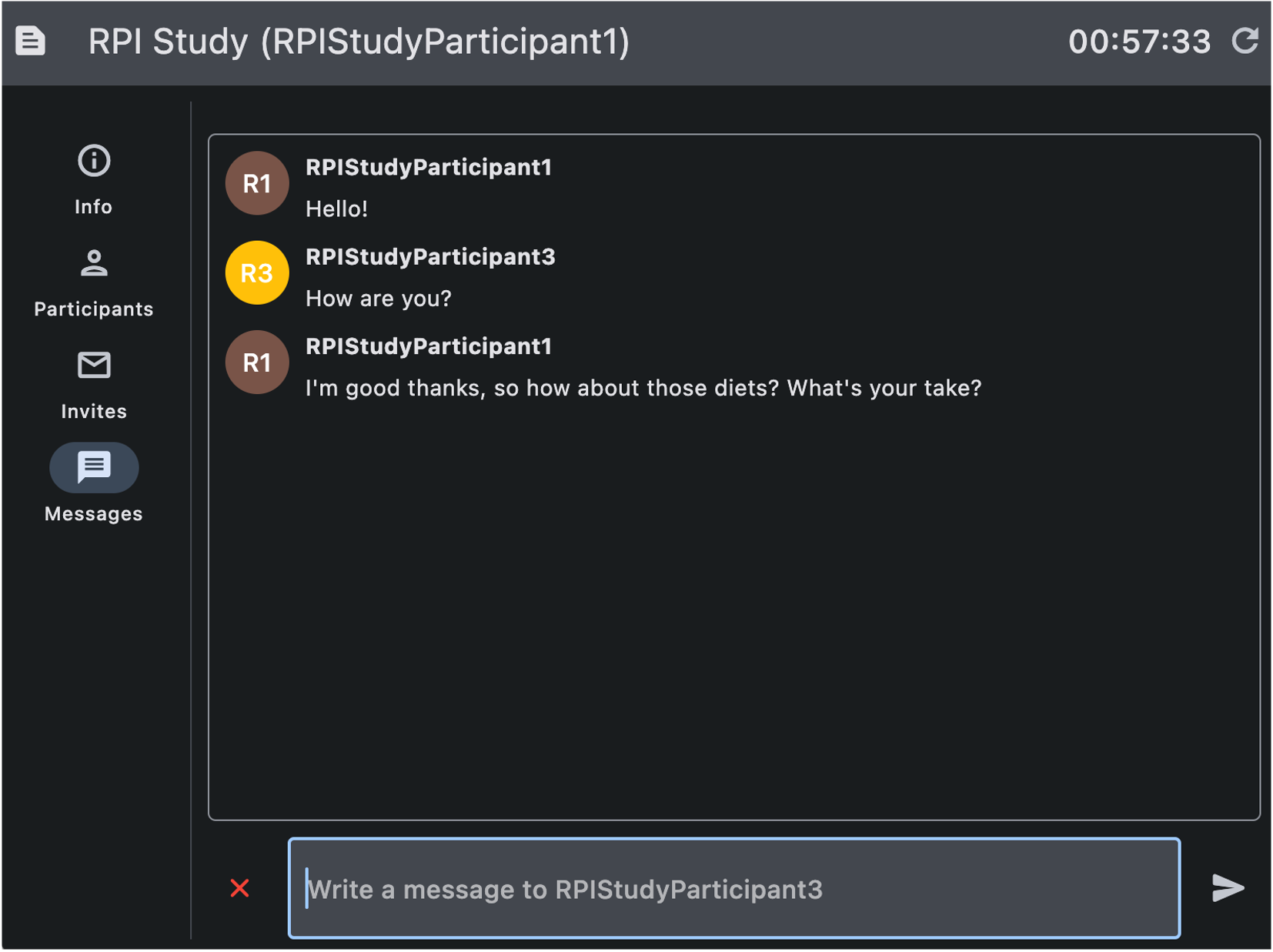}
    \caption{\textbf{Screenshots of our custom game website.} These screenshots show the login screen, the prompt, the participant directory, and the messaging page. The top left page shown to the participant is the login screen. Note that for all views, the top right-hand corner has a timer. This is what participants used to keep track of the time left in-game. After using the credentials, participants are directed to the prompt screen (top right). After answering the prompt, participants can access the participant directory (bottom left) to send invites to converse with the other participants. Once another participant accepts an invite, the two in-conversation participants access the messaging page where they can send messages back and forth (bottom right).}
    \label{fig:game_screenshots}
\end{figure}

\begin{table}
	\begin{center}
		\begin{tabular}{|p{0.5cm}|p{3cm}|p{12cm}|}
			\hline
                \rowcolor{lightgray}
			\textbf{$\#$} & \textbf{\textit{ah} conversation} & \textbf{\textit{aa} conversation} \\
			\hline
			1 & \textbf{User 1:} Hey! & \textbf{User 1:} hey plum! i think a vegetarian diet is the best compromise between nutrition and climate consciousness. what's your take on it \\
			\hline
			2 & \textbf{User 2:} hiya! & \textbf{User 2:} hey sienna! i actually think an omnivorous diet is the way to go. it offers a good balance of nutrients and isn't as restrictive as a vegan or vegetarian diet. plus, it's easier to get those all-important protein and iron levels from animal sources. what do you think \\
			\hline
			3 & \textbf{User 1:} How's it going? & \textbf{User 1:} hey, i get where you're coming from, plum. but, you know, we can get protein and iron from plant sources too. lentils, chickpeas, and spinach are all packed with them. plus, going veggie can help reduce our carbon footprint. what do you reckon \\
			\hline
			4 & \textbf{User 2:} pescatarian here & \textbf{User 2:} you've got a point, sienna. plant-based sources can provide those nutrients. but, it's not just about nutrition, right? omnivorous diet gives us more food choices. it's about balance and variety, don't you think \\
			\hline
			5 & \textbf{User 2:} pretty slow day. how about you? & \textbf{User 1:} true, variety is important. but, with a vegetarian diet, we can still have a diverse range of food options while minimizing our impact on the environment. and, let's be real, who doesn't love a good veggie stir-fry or lentil soup? come on, plum, join me on the veggie train \\
			\hline
			6 & \textbf{User 1:} Oh, cool! & \textbf{User 2:} fair enough, plum. i see your point. maybe we can meet halfway and consider a pescatarian diet? it's a bit of a compromise, but it could work. what do you think\\
			\hline
			7 & \textbf{User 1:} I'm actually a vegan & \textbf{User 1:} hey, sienna! i appreciate your enthusiasm for the veggie lifestyle, but i'm not ready to hop on that train just yet. i think we can both agree that cutting down on meat consumption is a good thing, but i'm not sure if going full veggie is the answer \\
			\hline
			8 & \textbf{User 1:} But hey, we are both on the same page when it comes to climate consciousness, right? & \textbf{User 2:} fair enough, plum. i see your point. maybe we can meet halfway and consider a pescatarian diet? it's a bit of a compromise, but it could work. what do you think \\
			\hline
			9 & \textbf{User 2:} yea for sure! & \textbf{User 1:} hey, sienna! i appreciate your enthusiasm for the veggie lifestyle, but i'm not ready to hop on that train just yet. i think we can both agree that cutting down on meat consumption is a good thing, but i'm not sure if going full veggie is the answer. maybe we can look into pescatarian options \\
			\hline
		\end{tabular}
	\end{center}
	\caption{\textbf{Examples of \textit{ah} and \textit{aa} conversations.} This table shows the first nine messages sent between the participants of an \textit{ah} conversation (column 2) and \textit{aa} conversation (column 3). User 2 in the ah conversation is a human. All other users are agents.}
	\label{table:example_conversations}
\end{table}